\documentclass[twocolumn,10pt]{technicalreport}

\def\BibTeX{{\rm B\kern-.05em{\sc ei\kern-.025em b}\kern-.08em
    T\kern-.1667em\lower.7ex\hbox{E}\kern-.125emX}}

\begin{document}
\title{Distributed Preemption Decisions: Probabilistic Graphical Model,  Algorithm and Near-Optimality}

\author{
\bf{Sung-eok Jeon and Chuanyi Ji }\\
sujeon@microsoft.com and jic@ece.gatech.edu\\
}

\maketitle
\begin{abstract}

Cooperative decision making is a vision of future network management and control. Distributed connection preemption is an important example where nodes can make intelligent decisions on allocating resources and controlling traffic flows for multi-class service networks. A challenge is that nodal decisions are spatially dependent as traffic flows trespass multiple nodes in a network. Hence the performance-complexity trade-off becomes important, i.e., how accurate decisions are versus how much information is exchanged among nodes. Connection preemption is known to be NP-complete. Centralized preemption is optimal but computationally intractable. Decentralized preemption is computationally efficient but may result in a poor performance. This work investigates distributed preemption where nodes decide whether and which flows to preempt using only local information exchange with neighbors.

In this work, we first model a large number of distributed preemption-decisions using a probabilistic graphical model. 
We then define the near-optimality of distributed preemption as its approximation to the optimal centralized preemption within a given error bound. 
We show that a sufficient condition for distributed preemption to be optimal is that local decisions should constitute a Markov Random Field. The decision variables, however, do not possess an exact spatial Markov dependence in reality due to the flows passing through multiple links. Hence we study traffic patterns of flows, and derive sufficient conditions on flows for the distributed preemption to be near-optimal. 
We develop, based on the probabilistic graphical models, a near-optimal distributed algorithm. The algorithm is used by each node to make collectively near-optimal preemption decisions. We study trade-offs between near-optimal performance and complexity that corresponds to the amount of information-exchange of the distributed algorithm. The algorithm is validated by both analysis and simulation. 
\end{abstract}


\section{Introduction}
\label{Introduction}

A vision of future network management is to involve nodes to make intelligent decisions on allocating resources and controlling traffic flows. This includes admitting new flows by preempting less important existing flows, which is well studied in the policy based admission control (i.e.,  admission is based on the priority of flows) \cite{Garay} \cite{JeonComComm}. Specifically, preemption is defined  at a prioritized multi-class network, where a new call needs  to be set up with a high priority between a source (S) and a destination  (D) \cite{Garay} \cite{JeonCommLetter} \cite{JeonComComm} \cite{Oliveira} \cite{Peyravian0}. When the capacity is insufficient at all feasible routes between the source-destination (S-D) pair, some existing flows of the lower priorities need to be forced to reduce their bandwidth, move to the lowest service class (e.g., best-effort-service), or simply preempted to accommodate the new call. Preemption decisions is to decide which lower priority flows to remove to free the reserved bandwidth for the new call at a chosen route \cite{Garay} \cite{Peyravian0}. The goal is to decide whether to preempt an active flow so that the total preempted bandwidth can be minimal under such constraints as bandwidth demand of a new call and available free bandwidth at each link \footnote{The preempted flows are usually rerouted to other paths. Hence preemption and rerouting can be considered jointly with somewhat different objectives \cite{Szv}. This work, however, focuses on preemption on a given path without considering rerouting.}.\\  
\indent
The benefit of preemption has been described in the prior works. For example, preemption allows a new high-priority connection to access heavily crowded core networks, e.g., multi-protocol label switched (MPLS) networks \cite{Szv}. Connection preemption also improves resource utilization by allowing low-priority flows to access unused bandwidths \cite{Herzog} \cite{Szv}. Preemption sees potential applications in emerging networks. For example, in 802.11e Wireless LAN, delay sensitive IP packets in expedited forwarding (EF) class can be served earlier than the best-effort packets through preemption \cite{Choi}. Multi-level preemption and precedence (MLPP) is proposed to classify calls by their importance, which can be used for military as well as commercial networks \cite{MLPP}.\\ 
\indent
There are two significant challenges for preemption which are performance and complexity. Performance corresponds to whether right flows are preempted to result in the minimal bandwidth to accommodate a new flow. Complexity corresponds to the amount of information needed for preemption decision. Preemption is known to be NP-complete \cite{Garay}. The complexity results from a large number of active flows supported by a core network for which preemption decisions need to be made. For example, for a 1Gbps link, if the bandwidth of each flow is in the order of Kbps, there would be thousands of flows supported per link.  In addition, a flow generally passes through multiple nodes, making preemption decisions among nodes dependent and thus difficult to be done with local information.  Thus preemption is network-centric, and may require a huge amount of information to perform in a large network.\\ 
\indent
For centralized preemption decisions, a centralized node maintains the routed-path information of active flows, their priorities and bandwidth occupancies at the entire route. The centralized node then decides which active flows to preempt upon the request of a new call. Therefore, centralized preemption can always be optimal, resulting in minimal preempted bandwidth. But the amount of management information needed can be overwhelming at the centralized node. 
For example, let $F_t$ be the total number of distinct flows per priority class at the route of a new call. Each flow has two states, preempted or not preempted. The total number of possible states is $O(2^{F_t})$ for making a centralized decision. When $F_t$ is in the order of hundreds or thousands \cite{JeonICC02}, centralized preemption becomes computationally intractable. Decentralized preemption is then adopted for reducing the amount of management information \cite{Peyravian0}.\\
\indent
Decentralized preemption is done at each node individually, and thus requires a node to maintain its local information, i.e., active flows at the adjacent links, their priorities and bandwidth occupancy. Such information is available locally at nodes. A node then decides, independently from the other nodes, which connections to preempt. This, however, may cause conflicting local decisions on the same flows that pass multiple links on the route, resulting in more preempted bandwidth than necessary.  In other words, decentralized preemption decision neglects the spatial dependence for the flows across multiple links, and may perform poorly. But the amount of management information are greatly reduced compared with centralized preemption. 

For example, let $F$ be the maximum number of active flows per link. The total number of states is $2^F$ at each link. 
Since $2^F$ $\ll$ $2^{F_t}$, compared with centralized preemption, decentralized schemes have a much smaller search space for preemption decisions. Therefore, most algorithms in the literature focus on decentralized preemption (see \cite{Oliveira} \cite{Peyravian0} and references there in).
\\
\indent
This work studies distributed decisions, that take into account spatial dependence among neighboring links through local information exchange. In fact, distributed preemption can be considered as a generalization of centralized and decentralized preemption. Centralized preemption corresponds to one extreme case of distributed preemption that an entire route is  the neighborhood for information exchange; whereas, decentralized decisions correspond to another extreme case where the neighborhood size is zero. Therefore, the communication complexity can be characterized in terms of neighborhood size. There is a trade-off between the optimality and the complexity. 
\\ 
\indent
In general, it has been shown to be a difficult problem to develop a distributed algorithm whose performance is predictable and within a tolerable degradation (i.e., given error bound) from that of the optimal scheme \cite{TSI}. 
Hence, the open issues are: (a) {\it When} can distributed decisions collectively result in a near-optimal global preemption? (b) {\it How} to model a large number of dependent decision variables and to obtain near-optimal local decisions using distributed algorithms? We apply machine learning to study these issues.\\

\indent
{\bf Machine learning perspective:} A machine learning view of distributed preemption is that individual nodes ``learn to make decisions" collectively and iteratively. Ideally, if each node has complete information on all active flows at the route of a new flow, the node will be able to make correct decisions on which flows to preempt. However, at any given time, a node has only partial information on the active flows on the route and its neighbors' decisions on the flows to preempt. But a node can adapt, i.e., learn to make decisions based on those of its neighbors'. As neighbors learn from neighbors' neighbors, a node would indirectly learn what farther nodes decide only with a delay. Eventually, all nodes would make local decisions, collectively resulting in a near-optimal preemption at the entire route.\\ 
\indent
How would machine learning benefit distributed preemption? The problem of collective learning and decision-making has been a keen interest in machine learning and adaptive control \cite{Baras} \cite{Ghosh}, but has just begun to see applications in networking. In particular, \cite{Doyle} proposes using Markov Random Fields as a general model of decision-making in Ad hoc wireless networks. The model is then applied to routing in wireless networks. Our prior work  \cite{JeonJi} \cite{Liu07} obtain probabilistic graphical models for ad hoc wireless and wireline networks starting from network properties \cite{Liu07}\cite{Liu08}, and the resulting probabilistic models turn out to be multi-layer. This work focuses on distributed decisions on network flows. We view machine learning as a framework in which a large number of decision variables can be treated jointly. Spatial dependence among these variables poses a key challenge to preemption, is an origin of high communication complexity,  and has not been dealt with sufficiently in prior works. Machine learning provides feasible approaches for this problem as summarized below.\\
\indent
(a) {\it Global model of distributed preemption decisions:} We first develop a probabilistic model that represents explicitly the spatial dependence of distributed preemption decisions over a pre-determined preempting route of a new flow. The randomness results from randomly arriving/departing active flows and their locations. The preemption decisions made on flows at each node are also random due to incomplete and inaccurate local information for distributed  preemption. We first obtain a cost function for preemption as a ``Hamiltonian" (or ``system potential energy") \cite{MRF}. A Hamiltonian combines local preemption decisions and constraints into a single quantity. The constraints include link capacity, unused bandwidths and bandwidth-demand of a new flow at each link. The Hamiltonian is then used to obtain a spatial probabilistic model as a Gibbs distribution \cite{Geman}.\\
\indent
(b) {\it Markov Random Field (MRF) and sufficient conditions:}  Spatial dependence can be characterized through a probabilistic dependency graph of graphical models \cite{Geman}\cite{Jordan}\cite{Kschischang} in machine learning. A probabilistic dependency graph provides a simple yet explicit representation of the spatial dependence among random variables. We show that if the dependence of decision variables is spatially Markovian, a globally optimal preemption decision can be obtained collectively by iterative local decisions through information exchange only with neighboring nodes. Such a probabilistic model is known as a Markov Random Field \cite{Geman}. 

In general, distributed decisions may not be spatially Markov, since the spatial dependence is caused by flows across multiple links. Hence we identify traffic patterns of active flows that result in approximately spatial Markov dependence. We then define the near-optimality of distributed decisions as the difference between the centralized and distributed decisions, measured in the Hamiltonian, and obtain sufficient conditions for the difference to reside within an error bound.\\
\indent
(c) {\it Distributed Decision Algorithm:} A near-optimal distributed algorithm is derived based on the Markov Random Field.  The algorithms can be implemented through either message passing \cite{Kschischang} or Gibbs sampling \cite{Geman}.\\
\indent
(d) {\it Trade-offs:} A challenging issue is the performance-complexity trade-off, i.e., ``when" and ``how" distributed preemption can achieve  a near-optimal performance with a moderate complexity. Here the {\it performance} measures the optimality of distributed preemption decision relative to that of the centralized optimal decision. The communication complexity of distributed preemption can be characterized by the amount of information used in distributed decision making. Distributed decisions reduce complexity using information exchange only with neighbors, but may deviate from the optimal performance.  Hence we study performance and complexity trade-off through both analysis and simulation.\\
\indent
The rest of this paper is organized as follows.
Section \ref{PS} provides a problem formulation on connection preemption.
Section \ref{Markov} develops a probabilistic spatial model of distributed preemption, utilizing the graphical models in machine learning and interpreting the derived model in terms of optimality and complexity. 
Section \ref{DA} proposes a distributed preemption algorithm based on the derived model, using probabilistic inference. 
Section \ref{PCs} analyzes the performance of distributed preemption. 
Section \ref{PE} validates the performance of distributed preemption through simulation. Section \ref{Priorwork} provides a further literature review and discussions. Section \ref{Conclusions} concludes the paper.

\section{Distributed Preemption}
\label{PS}

\subsection{Example}
\label{PS1}

\begin{figure} [htb] \centering
  \begin{tabular}{c}
    \resizebox{2.5in}{!}{\includegraphics{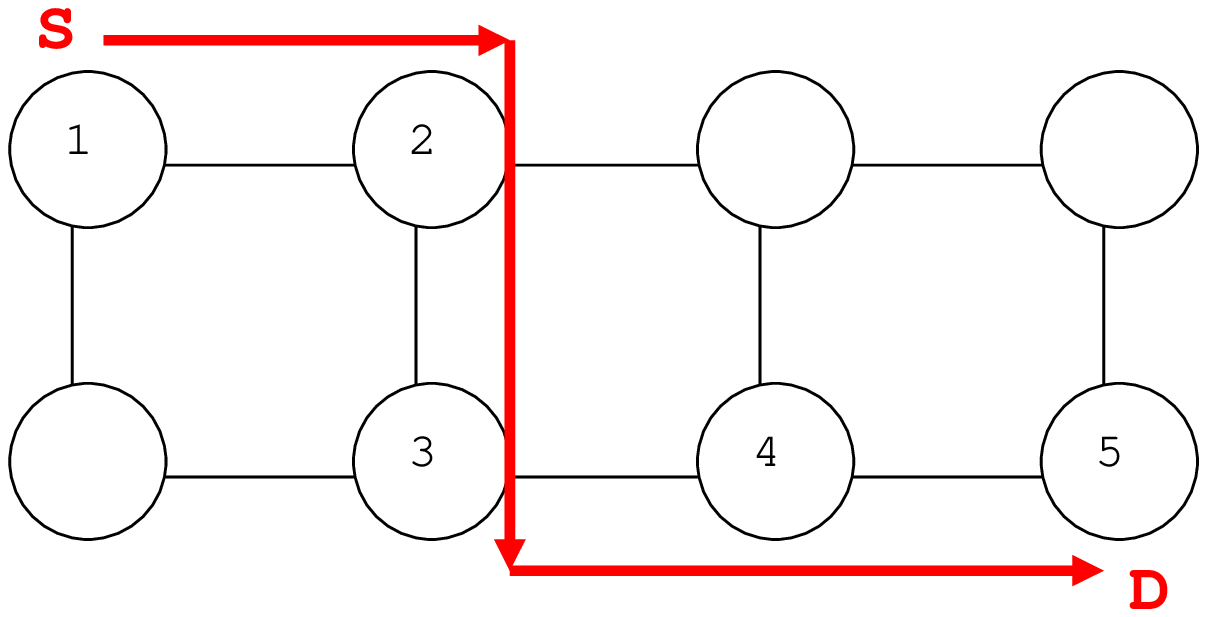}} \\ 
    {\footnotesize (a) A new call arrival at a network}\\
    \resizebox{3.2in}{!}{\includegraphics{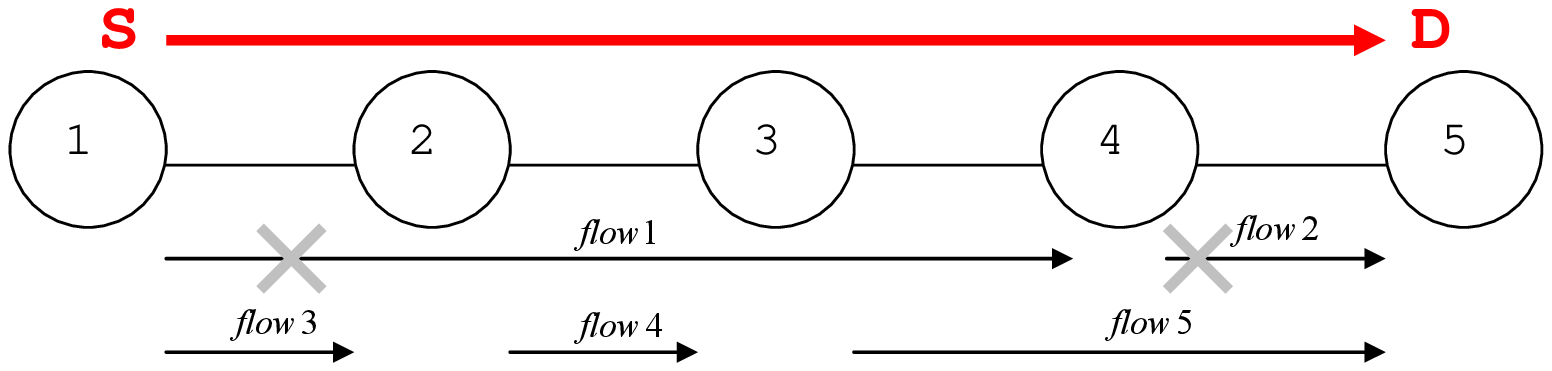}} \\
    {\footnotesize (b) Preemption decisions at each hop over the route of a new connection}
   \end{tabular}
\caption{Example of preemption} 
\label{Illustration}
\end{figure}

Figure \ref{Illustration} (a) shows an example network, and Figure \ref{Illustration} (b) illustrates distributed preemption on a given route.   
Assume that a new call request is made on the route between the SD pair, and for simplicity, all flows have the same bandwidth. 
To accommodate the new flow, the centralized preemption would preempt two existing flows, i.e.,  flow 1 and flow 2 that are marked with {\it X}. Such a preemption decision is obviously optimal. 
Now consider distributed preemption. In reality, distributed preemption decisions are made at nodes. For notational convenience, we regard the decision is made at the link which connects the left node.  For example, link 2 is the link beginning with node 2. The local information available at node 2 includes the priorities and the bandwidths of flows 1, 3 and 4 that pass through this node. When the bandwidth is the same for all flows, node 2 may decide to preempt flow 4 without knowing that nodes 1 and 3 both decide to preempt flow 1. Such a decision would result in more flows to be preempted than necessary compared with the centralized decisions. In contrast, node 2 would choose to preempt flow 1 if node 2 also has the information on the decisions made by the neighbors (nodes 1 and 3). This example shows the following facts.  

(a) Local decisions are spatially dependent. The spatial dependence originates from flows that trespass multiple nodes, and link capacity that constrains the aggregated flows on a link. 

(b) The spatial dependence can be taken into consideration by exchanging local decisions among neighbors. How many nodes should exchange local decisions depend on extents of flows. 

(c) The information exchanged would result in consistent decisions across a network, and thus improve the optimality of local preemption done at nodes.

These facts motivate cooperative distributed preemption formulated below.

\subsection{Problem Formulation}
\label{PF}

\indent
{\bf Assumptions}: We assume that a preempting route $R_p$
is pre-determined for a new connection \cite{Garay} \cite{Peyravian0}, and composed of node $1$, $\cdots$, and node $L+1$.  We assume that the traffic flows on the route belong to multiple priority classes $1$, $\cdots$, $i_{max}$, and a new connection belongs to class $i_{new}$ and demands bandwidth $c_{new}$.

{\bf Variables}: Let $S_F$ be a set of all active flows on route $R_p$, where $S_F$ = $\{ f^1,$ $\cdots,$ $f^{|S_F|} \}$ with $|S_F|$ being the cardinality of set $S_F$. $f^k$ and $B^k$ denote flow $k$ on the preempting route and its bandwidth. 
We consider the decisions at links. For notational convenience, without causing any confusion between nodes and links, we denote link $i$ as the link between node $i$ and node $i+1$ for simplicity for $1 \le i \le L$, where $L$ is the number of links on the considering route. 
Let $\mbox{\boldmath$f_{i}$}$ be the set of all active flows at link $i$ for $1 \le i \le L$. 

Let $d_{i}^k$ denote the preemption decision on flow $k$ at link $i$ for $1  \le i \le L$. $d_{i}^k$=1 if link $i$ decides to preempt the flow; $d_{i}^k$=0, otherwise. Let $\mbox{\boldmath$d_{i}$}$ denote the set of local preemption decisions on all active flows at link $i$. Let $\mbox{\boldmath$d$}$ denote all local decisions on the route, then $\mbox{\boldmath$d$}$ = $\{ d^k, \mbox{for} \; 1 \le k \le |S_F|$\}, where  $d^k$ denotes the preemption decision on flow $k$ over the entire preempting route,  
$d^k$ = $1 - \prod_{i=1}^L (1-d_{i}^k)$.   
Hence, $d^k=1$, i.e., flow $k$ is preempted from the given route, if at least one link decides to remove the flow, and $d^k=0$ if all links decide to keep the flow\footnote{Note that preemption in a general context can be considered as removing a high priority flow to the best-effort class rather than completely  terminating the service for the flow.}. Hence, $d^k=1$  is a global decision of the entire path, and a local decision $ d_{i}^k $ ($1 \le i \le L$ and $k \in \mbox{\boldmath$f_{i}$}$ ) impacts the global decision.

{\bf Problem statement}: Assume that the following information is maintained at node $i$ ($1 \le i \le L+1$): (a) complete local information on the active flows at link $i$ which includes flow ID $k$, class priority of flow $k$, bandwidth of flow $k$ $B^k$, for $k \in \mbox{\boldmath$f_{i}$}$ and  $1 \le k \le |S_F|$; and (b) neighbor 
information that includes decisions from the neighboring links within $N_d$ hops for $N_d \ge 1$. 

Given $\alpha_k$ ($\alpha_k >0$) as the priority weight\footnote{For example, if flow $k_1$ and $k_2$ belong to class $1$ and $2$ respectively, $\alpha_{k_1} < \alpha_{k_2}$.} of flow $k$, and $B_{i}^0 \ge 0$ as the amount of unused bandwidth at link $i$, for $1 \le i \le L$, preemption is to obtain a set of decisions $d_i^k$'s that 

\begin{eqnarray}
\label{obj0}
\mbox{minimize} 
& & \sum_{1 \le k \le |S_F|} \alpha_{k} B^{k} d^{k} \\ \nonumber
\mbox{subject to}& & \nonumber \\ \nonumber
& & c_{new} \le A_i, \nonumber \\ \nonumber
& & d_{i}^k \in \{0,1\},  \; \mbox{for} \; 1 \le i \le L, \nonumber  
\end{eqnarray}
\noindent
where $A_{i}$ = $\sum_k B^k d_{i}^k + B_{i}^0$ is the total available bandwidth at link $i$ for the new flow. The constraint requires that the sum of the unused bandwidth before preemption and the preempted bandwidth at any link $i$ should be sufficiently large for accommodating the new flow.  \\
\indent
 {\bf Goal}: The goal of this work is to approximate this global optimization problem through distributed preemption decisions. In particular, the first step is to derive a distributed algorithm that obtains a set of local preemption decisions made at links through information exchange with neighbors. The second step is to obtain near-optimality conditions under which distributed decisions approximate the globally optimal preemption decisions.\\ 
\indent
Note that the objective function in (\ref{obj0}) is the cost corresponding to the total preempted bandwidth. Such an objective function is used by most of the existing works \cite{Peyravian0} \cite{Garay} \cite{Stanisic}. Since    $d^k$ = $1 - \prod_{i=1}^L (1-d_{i}^k)$, a flow is preempted if at least one node decides to preempt the flow, i.e., $d_{i}^k=0$ for at least one $i$ and a given $k$. Hence,  the global optimization  in (\ref{obj0}) requires making consistent decisions on the same flow at all  links over the preempting route, i.e., $d_{i}^k$'s are all equal for a given $k$.

Distributed preemption is to determine $\{d^k_{i}\}$ for $1 \le i \le L$ and $k \in \mbox{\boldmath$f$}_{i}$ that minimize (\ref{obj0}) using the local and the neighbor information. Hence the objective of distributed preemption  is still global over the entire path. But the management-information exchange is local for making preemption decisions. Hence a key challenge is how to model and coordinate a large number of spatially-dependent local decisions to achieve the global objective of preemption in a fully distributed fashion.

Table 1 summarizes our notations.

\begin{table}[h!]
\caption{Notations} 
\centerline{
\begin{tabular}{|c|c| } \hline 
$d_i^k$ & Local preemption decision made at link $i$ for flow $k$ \\ \hline
$d^k$ & Global preemption decision of flow $k$ of the path \\ \hline
$B^k$ & Bandwidth of flow $k$ \\ \hline 
$L$ & The total number of links at the path\\ \hline 
\end{tabular}}
\label{Notations}
\end{table}

\section{Probabilistic Spatial Model of Preemption Decisions}
\label{Markov}

We begin by developing a global model to represent the spatial dependence of a large number of distributed preemption decisions. We then derive a local model as an approximation.  The global and local models are developed through probabilistic graphical models in machine learning.

\subsection{Global Model}
\label{Markov0}

A global model should include accurate spatial dependence resulting from flows, objectives and constraints on distributed preemption decisions.

\subsubsection{Deterministic Flows}

We first consider an example with a given set of active flows. Let $H(\mbox{\boldmath$d$})$ be the cost for setting up a new connection of a high priority. We express $H(\mbox{\boldmath$d$})$ by expanding the cost from (\ref{obj0}) and using the Lagrangian multiplier for the constraints,

{\small
\begin{eqnarray}
\label{obj}
 H(\mbox{\boldmath$d$}) 
&=& \sum_k \alpha_k B^k (\sum_{i_1} d_{i_1}^{k}-\\ \nonumber
& & \sum_{i_1} \sum_{i_2 \neq i_1} d_{i_1}^{k} d_{i_2}^{k} + 
 \sum_{i_1} \sum_{i_2 \neq i_1} \sum_{i_3 \neq i_1,i_2} d_{i_1}^{k} d_{i_2}^{k} d_{i_3}^k-\nonumber \\ \nonumber
& & \cdots \nonumber \\ \nonumber
& & +(-1)^{i_L} \sum_{i_1} \cdots \sum_{i_{L-1}} d_{i_1}^k \cdots d_{i_{L-1}}^k) \nonumber \\ \nonumber
& & + \beta  \sum_{i=1}^L U(c_{new}-A_{i}),
\end{eqnarray}
}
\noindent
where $\beta>0$ is a Lagrangian multiplier, $1 \le i_1 \neq i_2 \le L$, $1 \le k \le |S_F|$, $U(x)$ is an indicator function: $U(x)=0$ if $x \le 0$; $1$, otherwise. The $\beta$-term corresponds to the capacity constraint in (\ref{obj0}). 

Hence minimizing (\ref{obj}) corresponds to deterministic optimization (refer (\ref{obj0})), and conceptually, centralized preemption can always find an optimal set of flows to remove. 

Distributed preemption allows each link $i$ to update its decisions iteratively and asynchronously based on local information ($A_i$) and neighbors' decisions ($d_j^k$, $j \in N_d$, where $N_d$ is a neighborhood of link $i$). Each link/node can only access limited and initially inaccurate information from near-neighbors and missing information from far-neighbors. But through neighbor's neighbors, such information would eventually propagate to all nodes, resulting in globally consistent decisions. A difficulty is that deterministic distributed decisions may get stuck at a local optimum \cite{Geman}. 
%

\subsubsection{Random Flows} 

What and how many flows are active at which links are related to user behaviors and thus random. Hence active flows and their aggregation at individual links should be regarded as random variables. Preemption decisions made on active flows should be considered as random also. A set of decisions thus form a sample space $S_{\mbox{\boldmath$d$}}$ = \{$\mbox{\boldmath$d$}$\},  a subset of which consists of events due to distributed decisions. A given set of decisions on a given set of flows is then   
 a sample realization of an event. One such sample is shown in the example of Figure \ref{Illustration}, where $\mbox{\boldmath$d$}$=\{$d_{1}^1$, $d_{1}^3$, $d_{2}^1$, $d_{2}^4$, $d_{3}^1$, $d_{3}^5$, $d_{4}^2$, $d_{4}^5$ \}= \{1, 0, 1, 0, 1, 0, 1, 0\}. This relates random and deterministic flows and decisions. 

To obtain an optimal set of preempted flows, stochastic rather than deterministic optimization should be used, and this requires obtaining a probability distribution of $ \mbox{\boldmath$d$}$.\\ 
\indent
Such a probability distribution can be obtained through graphical models defined on neighborhood systems \cite{Geman}. A neighborhood system can be characterized by Hamiltonian which is also called system potential energy \cite{MRF}.  The energy of a decision variable corresponds to a per-variable preemption cost, $\alpha_{k} B^{k} d^{k}_{i_1}$, in the first terms of (\ref{obj}). Interactions between decision variables of any two different links result in $\sum_{i_1,k} \sum_{i_2 \neq i_1,k}$  $\alpha_{k} B^{k}$  $d_{i_1}^{k}  d_{i_2}^{k}$ as the second terms of (\ref{obj}). The remaining terms correspond to higher-order interactions.

In this context, $H(\mbox{\boldmath$d$})$ corresponds to a Hamiltonian of $\mbox{\boldmath$d$}$, and  results in a Gibbs distribution \cite{Geman} \cite{MRF},
{\small
\begin{eqnarray}
P(\mbox{\boldmath$d$}) &=& Z_0^{-1} \mbox{exp} \left ({-H(\mbox{\boldmath$d$}) \over T} \right ), 
\end{eqnarray}
}
\noindent
where $T$ is a parameter (the temperature \cite{Geman}), and $Z_0$ is a normalization constant. The Gibbs distribution is a probability distribution of decision variables, and thus provides a mathematical representation of the spatial dependence of distributed decisions. This Gibbs distribution also provides the implementation methodology of near-optimal distributed and iterative preemption decision, which is shown in Section \ref{DA}.
The minimum of the Hamiltonian corresponds to the optimal preemption decisions that maximize the probability.

\subsubsection{Probabilistic Graphical Models}

The  spatial dependence among a large number of decision variables can be represented explicitly by probabilistic graphical models. 
A graphical models relates a probability distribution of random variables
with a corresponding dependency graph \cite{Geman} \cite{Jordan} \cite{Kschischang}. 
A node in the graph represents a random variable and a link between two nodes characterizes their dependence.
In particular, a set of random variables $\mbox{\boldmath$v$}$ forms Gibbs Random Field (GRF) if it obeys a Gibbs distribution \cite{MRF}.
 A Gibbs distribution  satisfies the positivity condition, meaning that all decisions have a positive probability. One other important property is the spatial Markov dependence defined by the neighborhood system and shown by  Hammersley-Clifford theorem.\\

{\it Hammersley-Clifford Theorem\cite{MRF}:  
Let $\mbox{\boldmath$S$}$ be the set of nodes, $\mbox{\boldmath$S$} = \{1, \cdots, N\}$. Let $\mbox{\boldmath$v$}$ be a set of random variables,  $\mbox{\boldmath$v$}$ = $\{ v_1, \cdots, v_N \}$. 

$\mbox{\boldmath$v$}$ is said to be a Markov Random Field if (i) P($\mbox{\boldmath$v$}$) $>$ 0 for $\forall$ $\mbox{\boldmath$v$}$ in sample space; (ii) $P(v_i| \mbox{\boldmath$v$}_j \; \mbox{for} \; j \in \mbox{\boldmath$S$} \backslash \{i\})$ = $P(v_i| \mbox{\boldmath$v$}_j \; \mbox{for} \; j \in N_i)$, where $N_i$ is a neighborhood of node $i$ for $i \in \mbox{\boldmath$S$}$.

The random field $\mbox{\boldmath$v$}$ is also said to be a Gibbs Random Field if its probabilistic distribution can be written in the form P($\mbox{\boldmath$v$}$)=$\prod_{c \in \mbox{\boldmath$C$}} V_c(\mbox{\boldmath$v$} )$, where $c$ is a clique, $\mbox{\boldmath$C$}$ is the set of all feasible cliques, and $V_c(\mbox{\boldmath$v$})$ is a general positive function called a clique potential function.

There is an equivalence between a Gibbs Random Field and a Markov Random Field {\it if and only if} the Gibbs distribution P($\mbox{\boldmath$v$} $) possesses the spatial Markov property}.\\

\indent
Markov Random Fields correspond to an interesting type of probabilistic graphical models where a random variable is conditionally independent of the other nodes given its neighbors. The conditional independence is spatially nested, i.e., a node depends on its far neighbors through neighbors' neighbors. Such nested dependence can be observed explicitly through local connections among nodes in a dependency graph. The corresponding Gibbs distribution is thus factorizable over clique potentials \cite{Geman}. 

\begin{figure}[htb!]
\vspace{1 mm}
\epsfysize=1.5in
\centerline{\epsffile{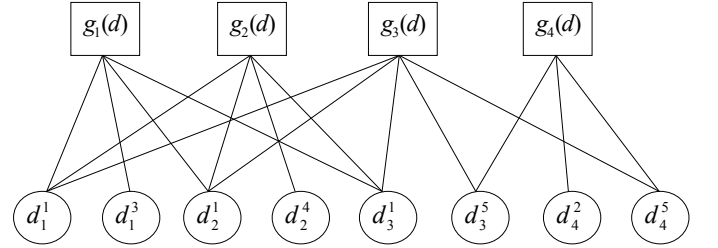}}
\caption{Spatial dependence of decision variables}
\label{FG1}
\end{figure}

An important implication to distributed preemption is that if distributed decisions result in an MRF, local decisions using neighbor information are collectively optimal.
But do preemption decisions $\mbox{\boldmath$d$}$ form a Markov Random Field in the first place? We plot the dependency graph for the Gibbs distribution with the Hamiltonian in (\ref{obj}). 
In particular, a factor graph \cite{Kschischang} in Figure \ref{FG1} is used to draw the dependency graph of the decision variables for the set of flows shown in Figure \ref{Illustration}. 

A factor graph is a bipartite graph that expresses the spatial dependence between the variable nodes and the function nodes \cite{Kschischang}. In Figure \ref{FG1}, circles represent the decisions on the active flows at links. Squares denote the link-functions, corresponding to the local potentials at individual links. Specifically, $g_{i}( \mbox{\boldmath$d$})$ is a local potential that encompasses the flows passing through link $i$. 
A connection between a circle and a square indicates a functional relation. Consider link $1$ in Figure \ref{Illustration} as an example. Flow $1$ passes link $1$ and extends to link $2$ and $3$. 
Multiple variables relating to the same local potential are dependent, e.g., $d_{1}^1$ and $d_{4}^5$ are dependent through $g_{3}(\mbox{\boldmath$d$})$. This is because flow 1 and flow 5 pass the same link $3$ and are thus dependent due to the limited capacity constraint. Meanwhile, different local potentials can be dependent if they share some flow-variables, e.g., $g_{1}(\mbox{\boldmath$d$})$ and $g_{3}(\mbox{\boldmath$d$})$ are both connected to $d_{1}^1$ and $d_{3}^1$. This is because flow 1 passes link $1$ and link $3$. This shows the global dependence, resulting from long flows which extend to far neighboring links/nodes.

Such statistical dependence can be represented quantitatively by local potentials $\sum_{i=1}^L g_{i}({\bf d})=H({\bf d})$ as in (\ref{obj}), for example, $g_{1}( \mbox{\boldmath$d$})$ in Figure \ref{FG1} is a local potential that encompasses the flows passing through link $1$, i.e., 
{\small
\begin{eqnarray}
\label{g_ij}
g_{1}( \mbox{\boldmath$d$}) &=& 
\alpha_{1} B^1 \left ( d_{1}^1-d_{1}^1d_{2}^1-d_{1}^1d_{3}^1 + d_{1}^1d_{2}^1d_{3}^1 \right ) + \alpha_3 B^3 d_{1}^3 \nonumber \\ \nonumber 
& & + \beta U(c_{new}-A_{1} ),\nonumber
\end{eqnarray}
}
\noindent
where $\alpha_1$=$\alpha_3$ because both flows $1$ and $3$ belong to class $1$.

Hence the graphical and the mathematical representations show that in general a decision random variable at a flow (circle on the graph) can have connections with both near and far neighboring local potentials (squares on the graph). This implies that in general, the decision variables are not spatially Markov, and the Gibbs distribution is thus not factorizable.   

\subsection{Local Model}
\label{Markov2}

If the long-range spatial dependence can be removed from the probabilistic dependency graph, the spatial dependence can be approximated through a spatial Markov model, i.e., a Markov Random Field. Such a Markov Random Field considers only dependence of decision variables with their neighbors, resulting in a truncated Hamiltonian as follows,  
{\small
\begin{eqnarray}
\label{localmodel}
H^l(\mbox{\boldmath$d$}) 
&=& \sum_k \alpha_k B^k (\sum_{i_1} d_{i_1}^{k} - 
 \sum_{i_1} \sum_{|i_2-i_1| <= N_d} d_{i_1}^{k} d_{i_2}^{k} )  \nonumber \\ \nonumber
& & + \beta \sum_{i=1}^L U(c_{new}-A_{i} ),
\end{eqnarray}
}
\noindent
where $1 \le i_1 \neq i_2 \le L$, $1 \le k \le |S_F|$, and $N_d$ denotes the neighborhood size of a node.\\
\indent
The corresponding Gibbs distribution is
{\small
\begin{equation}
P^l(\mbox{\boldmath$d$}) = Z_0^{-1} \exp \left ({-H^l(\mbox{\boldmath$d$}) \over T}\right ).
\end{equation}
}

\noindent
$P^l(\mbox{\boldmath$d$})$ is an approximated likelihood function, 
{\small
\begin{equation}
P^l(\mbox{\boldmath$d$}) = Z_0^{-1} \prod\limits_{i=1}^L \exp \left ({-g_{i}(\mbox{\boldmath$d$}) \over T} \right ),
\end{equation}
}
\noindent
where $\exp \left ({-g_{i}(\mbox{\boldmath$d$}) \over T} \right )$ is a local likelihood function for the connections at link $i$, and can be further decomposed into all clique potentials associated with connections at link $i$: 
{\small
\begin{eqnarray}
\exp \left ({-g_{i}( \mbox{\boldmath$d$}) \over T} \right)
&=& \mbox{exp} \left ({- \sum_{c \in C_{i}} \psi_c(\mbox{\boldmath$d$}) \over T} \right),
\end{eqnarray}
}
\noindent
where $C_{i}$ is the set of all cliques of link $i$, and $\psi_c(\mbox{\boldmath$d$})$ is a potential function of  clique $c$.

\begin{figure}[htb!]
\vspace{1 mm}
\epsfysize=1.4in
\centerline{\epsffile{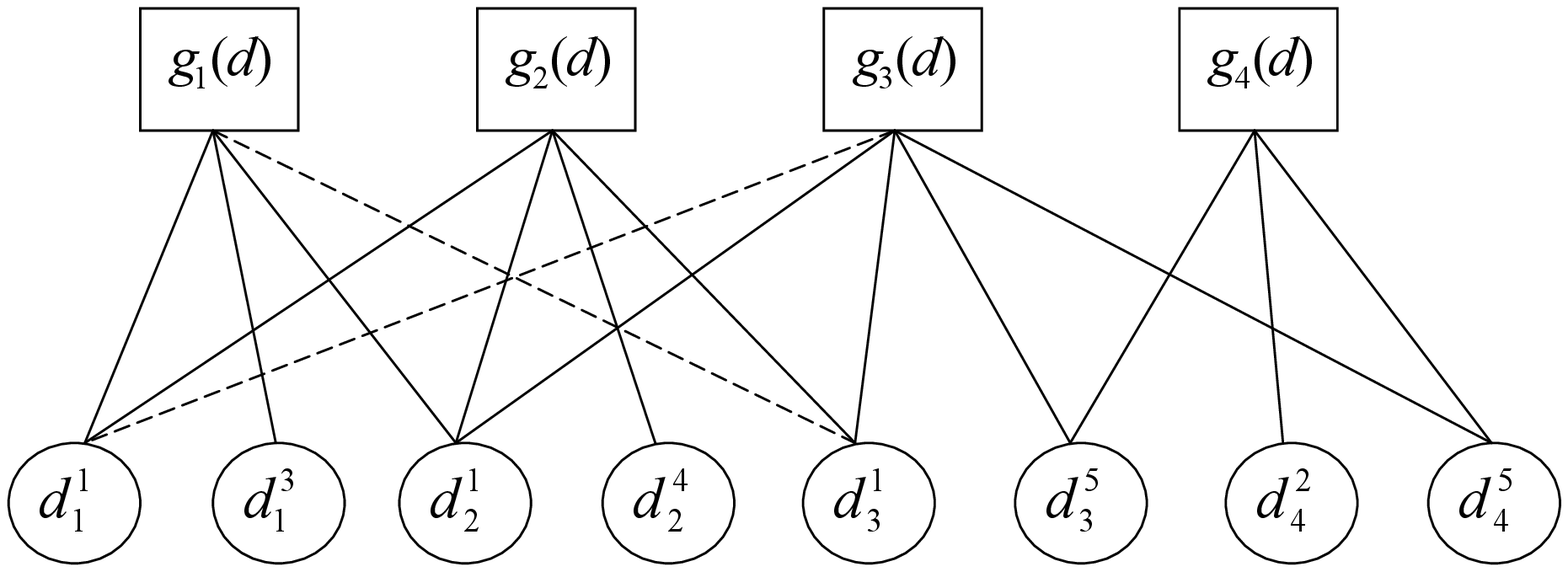}}
\caption{Localized spatial dependence of $ \mbox{\boldmath$d$}$ with Factor Graph}
\label{FG_local}
\end{figure}

For example, if the neighborhood size $N_d=1$ for all links, the corresponding factor graph has only nearest neighbor connections as shown in Figure \ref{FG_local}, where the dash lines denote the neglected dependency links.

%

\section{Distributed Preemption Algorithms}
\label{DA}

 We now assume that a local model is obtained as a good approximation of the global model. The spatial Markov local model then can be used to derive a distributed algorithm 
where nodes can make local decisions on connection preemptions through information exchange with neighbors.

\subsection{Distributed Algorithm}
\label{SO}

The distributed algorithm obtains a set of local decisions that maximizes the approximated likelihood function, which is equivalent to minimizing the cost function, 
{\normalsize
\begin{eqnarray}
\mbox{\boldmath$\hat{d}$}
&=& \mbox{arg} \mathop{\mbox{max}}_{\small \mbox{\boldmath$d$}} P^l(\mbox{\boldmath$d$}) \nonumber \\ 
&=& \mbox{arg} \mathop{\mbox{min}}_{\small \mbox{\boldmath$d$}} H^l(\mbox{\boldmath$d$}).
\end{eqnarray}
}
\indent
Since $P^l(\mbox{\boldmath$d$})$ is factorizable, 
maximizing the global likelihood function reduces to maximizing the local likelihood function at cliques, i.e., 
{\small $P^l(\mbox{\boldmath$d_{i}$} | \mbox{\boldmath$d_{N_{i}}$})$} for $1 \le i \le L$, where $\mbox{\boldmath$d_{N_{i}}$}$ is the set of decision variables of neighboring links. 
As these local likelihoods are functions of the decision variables of neighboring links, the decisions can be updated locally.
In addition, the local maximizations result in coupled equations due to the nested Markov dependence, which shows that information exchange is needed only among neighbors.\\
\indent
Maximizing local likelihood functions can be implemented as local learning algorithms at individual nodes. The learning algorithms perform probabilistic inference using either approximated sum product algorithm \cite{Kschischang} or stochastic relaxation \cite{Geman}. The sum product algorithm can be applied to the factor graph in Figure \ref{FG_local}. This algorithm produces an exact solution for a graph that has no loops.  However, the factor graph \cite{Kschischang} of preemption problem is usually loopy, resulting in approximated (non-optimal) decisions.

Stochastic relaxation can be applied for each link to make local preemption decisions. Let $\mbox{\boldmath$d_{i}^{S_F \setminus \{k\}}$}$ be a set of decisions on active flows at link $i$, excluding the decision on flow $k$. Here $S_F \setminus \{k\}$ denotes a set operation, which excludes $k$ from $S_F$.
Now we add time variable $t$ to the decisions \footnote{Distributed decisions depend on the iterative and cooperative decisions.}, and let $d_{i}^k(t+1)$ be an updated decision on flow $f^k$ at the $(t+1)$th iteration and at link $i$. Then, 
{\small
\begin{equation}
\label{table1}
d_{i}^k(t+1) =1,
\end{equation}
}
\noindent
with the probability  
{\small
\begin{eqnarray}
& & P^l \left (d_{i}^k(t+1) = 1 | \mbox{\boldmath$d_{i}^{S \setminus \{k\}}$}(t), \mbox{\boldmath$d_{N_{i}}$}(t) \right )=  \nonumber \\ \nonumber
& & \nonumber \\ \nonumber
& & {\mbox{exp} \left (-\psi_{i}(d_{i}^k(t+1)=1)/T(t+1) \right ) \over \sum_{d_{i}^k(t+1) \in \{-1,1\}} \mbox{exp} \left (-\psi_{i}(d_{i}^k(t+1))/T(t+1) \right )},  \nonumber
\end{eqnarray}
}
\noindent
where
{\small
\begin{eqnarray}
\label{psi_ij}
\psi_{i} \left (d_{i}^k(t+1) =1 \right ) &=& \alpha_k B^k -  
\sum_{i_2 \in N_{i}} \alpha_{k} B^{k} d_{i_2}^{k}(t+1)+ \nonumber \\ \nonumber
& & \beta  U(c_{new}-A_{i}). \nonumber
\end{eqnarray}
}

\indent
This means that a random decision at time epoch $t+1$ is made based on local information $\mbox{\boldmath$d_{i}^{S_F \setminus \{k\}}$}(t)$ and neighbor information $\mbox{\boldmath$d_{N_{i}}$}(t)$ at the previous time epoch $t$. A cooling schedule is applied to the temperature $T(t)$ = $T_0 / \mbox{log}(1 + t)$ with $T_0$=$3.0$. This results in an almost-sure convergence of the algorithm to the global minimum Hamiltonian (i.e., optimal decisions) \cite{Geman}. 
That is, with the iterative and distributed updates, the global minimum of the approximated Hamiltonian $H^{l}( \mbox{\boldmath$d$})$ can be reached asymptotically with probability one.

\subsection{Example}

We now revisit Figure \ref{Illustration} to show an  example of the distributed algorithm. Consider links $1$ and $2$, and assume that the neighborhood size $N_d$=1. That is, a node only exchanges information with its nearest neighbors. 

At initial stage, no flows are preempted, i.e. \{$d_{1}^{1}(0)$=0, $d_{1}^{3}(0)$=0, $d_{2}^1(0)$=0, $d_{2}^4(0)$=0\}. When $t=1$, the decision variables are updated, 
\begin{eqnarray}
\label{M1}
d_{1}^1(1) &=& \mbox{arg} \mathop{\mbox{max}}_{d \in \{0,1\}} \; 
P(d_{1}^1(1)=d| d_{1}^3(0), d_{N_{1}}(0)),  \nonumber
\end{eqnarray} 
\noindent
where $d_{N_{1}}(0)$ = \{$d_{2}^1(0), d_{2}^4(0)$\}. The updated decision $d_{1}^1(1)$ is sent to the neighboring links. This process is applied similarly to the other decision variables. At the second time epoch ($t=2$),
\begin{eqnarray}
\label{M2}
d_{2}^4(2) &=& \mbox{arg} \mathop{\mbox{max}}_{d \in \{0,1\}} \; 
P(d_{2}^4(2)=d| d_{2}^1(1), d_{N_{2}}(1)), \nonumber
\end{eqnarray} 
\noindent
where $d_{N_{2}}(1)$ = \{ $d_{1}^1(1)$, $d_{1}^3(1)$, $d_{3}^1(1)$, $d_{3}^5(1)$\}. 

The process is repeated until an equilibrium state (i.e., of no more changes) is reached.

\subsection{Information Exchange}

The distributed preemption decisions require information exchange with neighbors.  The clique structure of the Markov Random Field determines the range of information exchange, which is the neighborhood size $N_d$. The type of the information exchanged is binary, i.e., $\mbox{\boldmath$d_{N_{i}}$}(t)$, as in the conditional probability in (\ref{table1}). The amount of information used at a decision making characterizes the communication/computation complexity. The information exchange is per-flow based but moderate when limited to neighbors.

\section{Near-Optimality and Complexity}
\label{PCs}

In this section, we conduct analytical studies to identify sufficient conditions for the near-optimality of the distributed preemption, the communication/computation complexity, and the optimality-complexity trade-off.

\subsection{Short-Range Dependent Decision Variables}
\label{Markov1} 

The near-optimality is in regard to the question when distributed preemption decisions are nearly optimal. To answer this question, we need to consider how well a Markov Random Field approximates the global model. This should be done by studying the traffic patterns of active flows since the flows across multiple links over the preempting route and the limited link-capacity constraints at links are the origins of spatial dependence of distributed decisions.

\subsubsection{Bounded-Length Flows}

Traffic patterns of active flows result in spatial dependence among distributed decision variables. Consider simplified traffic patterns where the hop-count of each active flow is bounded by $h$ for $h \ge 1$. Then the set of distributed preemption decision variables are strictly Markov as shown below.\\

Lemma 1: {\it Assume that the hop-count of each active flow is bounded by $h$ ($h \ge 1$). Let $N_{i}^h$ be a set of neighborhood of link $i$ and include all links within $h$ hops from ($i,j$). Let $\mbox{\boldmath$d_{{}_{N_{i}^h}}$}$=$\{\mbox{\boldmath$d_{m}$}$, for $\forall m\in N_{i}^h\}$ denote a set of decisions in the neighborhood, and $\mbox{\boldmath$d_{{}_{}}$} \backslash \mbox{\boldmath$d_{i}$} $ be all decision variables except $\mbox{\boldmath$d_{i}$}$. Then, $P( \mbox{\boldmath$d_{i}$} | \mbox{\boldmath$d_{{}_{}}$} \backslash \mbox{\boldmath$d_{i}$} )$ $=$ $P( \mbox{\boldmath$d_{i}$} | \mbox{\boldmath$d_{{}_{N_{i}^h}}$} )$.\\ 
}

\indent
The proof is provided in Appendix \ref{Lemma1}. {\it Lemma  1} shows that the set of decision variables on active flows of a limited span forms a Markov Random Field (MRF), where $h$ corresponds to an upper bound of the neighborhood size of the MRF. This is intuitive as the active flows of a bounded length would only introduce short-range spatial dependence.

\subsubsection{Shortest-Path Flows}

In reality, however, the hop-count of active flows is a variable and cannot be assumed to be bounded with a meaningfully small value (e.g., $1$ or $2$ hops). Thus, we study the spatial dependence of decision variables for shortest-path flows that constitute more realistic traffic patterns. In particular, we consider shortest-path flows with the following assumptions for analytical convenience: 

(1) A network is planar and homogeneous where each node (except edge nodes) has the same nodal degree $d_0$ ($d_0 \ge 2$). 

(2) A source-destination pair is chosen randomly from all pairs in the network.

(3) A preempting route is a shortest-path between the source and the destination of a new connection. 

(4) Active flows are assumed to take shortest routes from randomly-chosen source-destination pairs whose paths may partially coincide with the route of the new connection.

(5) The route length of a new flow is assumed to be $L$ hops for analytical simplicity, where $L \ge 1$.\\

\indent
We now define a measure of spatial dependence of two links on the preempting route (e.g., link ($i-1$,$i$) and ($j$,$j+1$) for $2 \le i$, and  $i \le j \le L$).\\

\indent
Definition 1. {\it Link-Dependency Probability $P_{ij}$: $P_{ij}$ denotes the probability that a flow uses both link ($i-1,i$) and ($j,j+1$) that are separated by $|j-i|$ hops on the preempting route.}\\

\indent 
The link-dependency probability $P_{ij}$ then characterizes the spatial dependence of any two flows at these two links. $P_{ij}$ is difficult to obtain exactly, and thus bounded as follows.\\  

\noindent
Lemma 2: {\it Let $P^l_{ij}$ be a lower bound of $P_{ij}$, i.e., $P_{ij} \ge P_{ij}^l$. For shortest-path flows under assumptions (1) through  (5), $P^l_{ij}$ = $({L-|j-i| \over L})({1 \over d_0-1})^{|j-i|}$.\\
}

\indent
The proof is provided in Appendix \ref{Lemma2}. This lemma suggests that the length of a shortest-path flow over the preempting route follows at least a geometric probability, where ${1\over {d_0-1}}$ is the lower bound of the  probability for such a flow to continue at the next hop.\\  

\noindent
Lemma 3: {\it Let $P^u_{ij}$ be an upper bound of $P_{ij}$, i.e., $P_{ij} \le P_{ij}^u$. Consider a network topology of a regular lattice with nodal  degree $4$. For shortest-path flows under assumptions (1) through  (5), 
{\small
\begin{eqnarray}
P_{ij}^u &=&
\left \{
\begin{array}{ll}
({L-|j-i| \over L}) { C \left (|j-i|, {|j-i| \over 2} \right) \over 2 (2^{|j-i|}-1) }, & |j-i|=2 \\
({L-|j-i| \over L}) { C \left (|j-i|, {|j-i| \over 2} \right) \over 3 (2^{|j-i|}-1) }, & |j-i|>2,
\end{array}
\right.
\end{eqnarray}
}
\noindent
where $C(a,b)$=${a! \over (a-b)! b!}$ is a combinatorial coefficient, and  $|j-i| \ge 2$. For $|j-i| >> 1$ , $P_{ij}^u$ $\approx$ $({L-|j-i| \over L}){1 \over 3 \sqrt{2 \pi |j-i|}}$.\\
}

\indent
The proofs can be obtained by counting the number of shortest paths between node $i$ and $j$, and is given in Appendix \ref{Lemma3}. 

Figure \ref{Pu_ij} depicts both the upper and lower bound as well as an  empirical probability $P_{ij}$. The probability $P_{ij}$ is estimated on a regular lattice network with 250 nodes, where active flows are routed onto the shortest paths between randomly chosen source-destination pairs. $10$ runs are conducted and the results are averaged to obtain the empirical probability.  
As shown in the figure, $P_{ij}$ decays rapidly close to the exponential decreasing rate of the lower bound $P_{ij}^l$.

{\it Lemma} 2, 3 and the empirical result suggest that on the average, shortest-path flows share only few hops with the preempting route. Thus, as we shall soon see, Markov Random Fields is a good approximation to a set of decision variables $\mbox{\boldmath$d$}$ for a mesh topology.

\begin{figure}[htb!]
\epsfysize=3.0in
\centerline{\epsffile{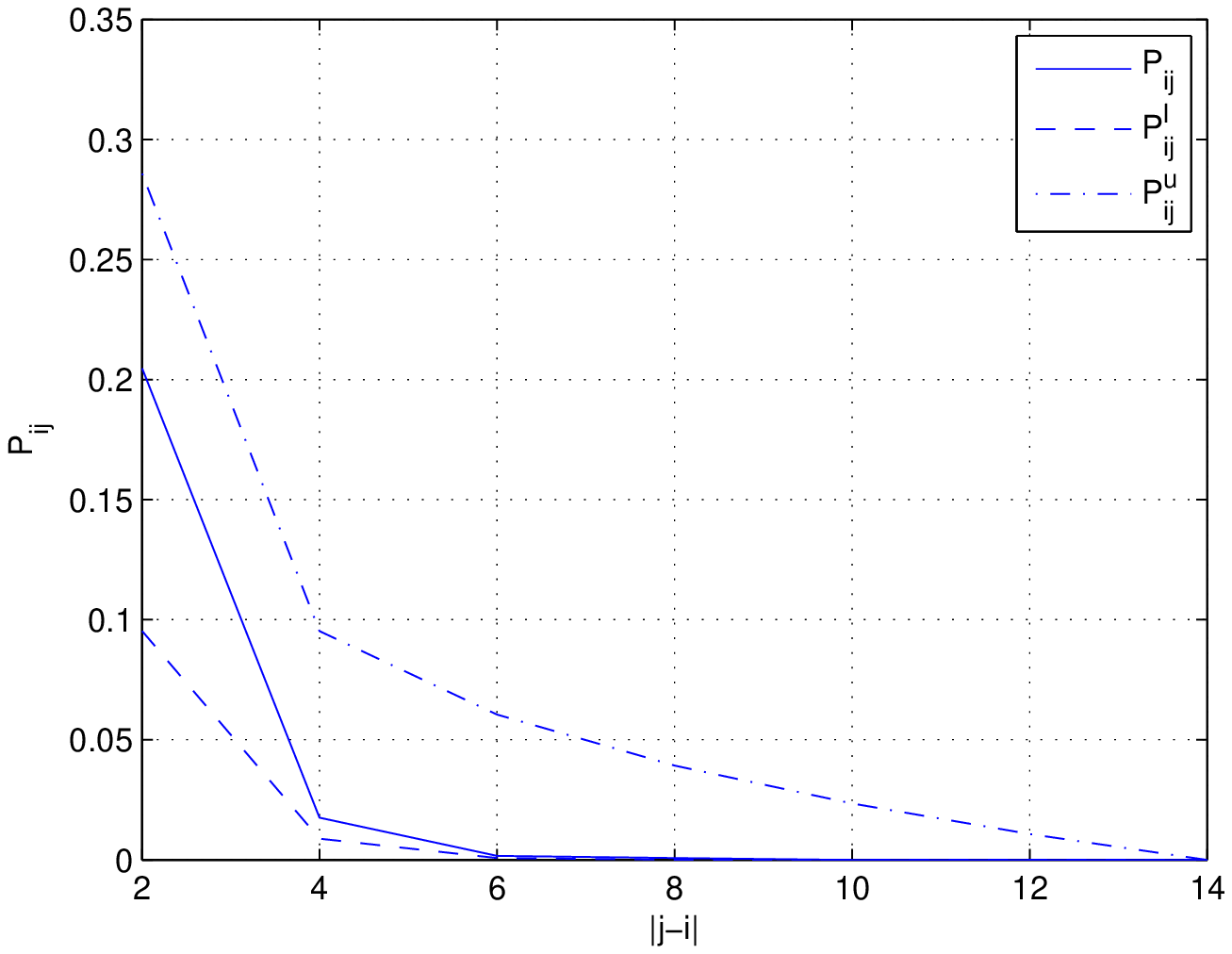}}
\caption{Upper and lower bounds of the probability that a flow visits both links ($i-1$,$i$) and ($j$,$j+1$) on the preempting route}
\label{Pu_ij}
\end{figure}
%


\subsection{Sufficient Conditions for Near-Optimality}
\label{LocalGlobal}

We now define the near-optimality of distributed preemption.\\

\indent
Definition 2. {\it Near-optimality of distributed decisions:
Consider a given route of a S-D pair of a new connection. Consider random flows on the route that obey $P_{ij}$, i.e., a flow would continue at $|j-i|$ hops from the source with probability $P_{ij}$. 
Let $ \mbox{\boldmath$d^*$}$ and $ \mbox{\boldmath$\hat{d}$}$ be two sets of preemption decisions that minimize the global Hamiltonian $H()$ and its approximation $H^l()$, respectively.
The optimality of distributed decisions $ \mbox{\boldmath$\hat{d}$}$ is measured by the expected value of the difference, i.e., $E(\Delta)$, where 
{\small $\Delta$ = $|H(\mbox{\boldmath$d^*$}) - H(\mbox{\boldmath$\hat{d}$})|$}. 
The expectation is over random flows and randomized decisions. 
 
Given a desired performance $\epsilon > 0$, if $E(\Delta) \le \epsilon$, $\mbox{\boldmath$\hat{d}$}$ is near-optimal.}\\

\indent
We now derive sufficient conditions for the near-optimality. 
This suffices to investigate whether and when the long-range dependence of active flows can be neglected in the global Hamiltonian. For feasibility of analysis, we consider the traffic patterns with a geometric probability drawn from {\it Lemma} 2 and 3. Such traffic patterns exhibit a certain practical relevance as shown in the above section, and is also analytically tractable.\\ 

\indent
Definition 3. {\it Flow-continuity probability $p_c$ is the probability that an active flow continues onto the next link on the preempting route.}\\

\indent
The notion of flow-continuity probability $p_c$ has been used in two other contexts to describe the extent of an optical flow \cite{Barry} \cite{Liu}. As shown in Lemma 3, $p_c$ characterizes the range of dependence of active flows, and corresponds to a special case of $P_{ij}$. In fact, if a flow continues with probability $p_c$ at each link independent of the other links, the length of an active flow would obey a geometric probability \cite{Liu}. For example, $p_c={1\over {d_0}-1}$ for the lower bound of the continuity probability of the shortest-path flows over a planar lattice topology with node degree $d_0$. 

Even when a large number of flows are short-range dependent, there can still be long flows. So a sufficient condition of the near-optimality needs to specify when the effects of aggregated long flows are negligible in the truncated Hamiltonian. For feasibility of analysis, we consider a simplified scenario that the bandwidths of active flows are bounded.\\ 

\indent
Theorem 1: {\it Let $B_0>0$ be a constant bandwidth. Consider a straight route of a new flow. For given $\epsilon_B$ ($0< \epsilon_B < 1$), assume that bandwidth $B^k$ of flow $k$ satisfies $|{B^k -  B_0 \over B_0}|$ $\le$ $\epsilon_{B}$  for all $k$. Then
$E(\Delta)$ $\le$ $2c_{new} \cdot {1+\epsilon_{B} \over 1 - \epsilon_{B}} \cdot L \cdot [(1+p_c^{N_d})^{L-N_d}-1]$, 
where $N_d$ is the neighborhood size for information exchange in distributed preemption decisions. 
When $p_c^{N_d}L = o(1)$, the upper bound of $2c_{new} \cdot {1+\epsilon_{B} \over 1 - \epsilon_{B}} \cdot L \cdot [(1+p_c^{N_d})^{L-N_d}-1]$ = 
$2c_{new} \cdot {1+\epsilon_{B} \over 1 - \epsilon_{B}} \cdot L \cdot (L-N_d)p_c^{N_d} +o(L p_c^{N_d})$.\\  
}

\indent
The proof is provided in Appendix \ref{Theorem1}. Theorem 1 provides the following observations when active flows follow a geometric distribution. 

(a)  For a given $p_c$ and $c_{new}$, the larger the neighborhood size $N_d$ in the Markov Random Field, the smaller the upper bound, and the better the performance may be for distributed preemption.  In fact, the error bound decreases exponentially with respect to $N_d$. 

(b) The upper bound increases linearly with respect to the bandwidth demand of a new flow $c_{new}$ as ${c_{new} \over {B_0} (1-\epsilon_B)}$ characterizes the maximum number of active flows to accommodate the new flow with $c_{new}$ at a link. Thus, 
the larger the $c_{new}$, the more existing flows may need to be preempted, the higher the probability for the distributed algorithm to make inconsistent decisions at links. That is, the performance of distributed preemption may degrade when the bandwidth demand of a new flow increases.

(c) The upper bound also increases with respect to the route length $L$, since a longer route consists of more links and thus a higher probability for distributed decisions made at links to be inconsistent.

It should be noted that the above studies of the optimality assume that the stochastic relaxation is capable of obtaining a global minimum of the global and local models. This holds true as the convergence of the algorithm occurs almost surely \cite{Geman}.

\subsection{Complexity}
\label{Complexity}

A key advantage of distributed preemption is the reduced complexity, i.e., the information exchange is limited to only neighbors.\\ 

\indent
Definition 4. {\it Communication Complexity (CC): Let $N_d$ denote the neighborhood size for exchanging binary information (bits) in distributed preemption. Let $f_{max}$ denote the maximum number of active flows at a link. Let $i_{ter}$ denote the total number of iterations needed for the distributed algorithm to converge. Communication complexity (CC) of a node is defined as the total amount of information exchanged for a link to make a decision using the distributed algorithm, i.e., $CC$ = $O(N_d f_{max} i_{ter})$. }\\

\indent
Note that $O(N_d i_{ter})$ is the bits of information exchanged for making a preemption decision on one flow. There are at most $f_{max}$ flows at a node. This results in  $CC=O(N_d f_{max} i_{ter})$. Hence, if $i_{ter}$ can be bounded by a moderate value, as shall be shown in Section VI, $CC$ would be $O(N_d f_{max})$ which grows linearly with respect to the neighborhood size and the number of active flows. 
 
We now compare qualitatively $CC$ with decentralized preemptions. Min-Conn and Min-BW \cite{Peyravian0} are the representatives of the  existing decentralized algorithms that minimize the number of preempted flows and the amount of preempted bandwidth at each hop, respectively, without information exchange. The complexity of Min-Conn and Min-BW are $O(f_{max}^2)$ and $O(f_{max}2^{f_{max}})$, respectively. 

Hence by bounding $i_{ter}$ with a moderate value, we can obtain a globally near-optimal decision that is obtained with a smaller complexity than that of decentralized algorithms. We shall show this in the next section.\\

We now compare qualitatively $CC$ with centralized preemptions. The communication/computation complexity for a centralized scheme increases linearly with the number of hops on the preempting route $L$. Hence, if $N_d << L$, the complexity for distributed preemption is much smaller than that of the centralized preemption due to local information exchange among neighbors.

\subsection{Optimality and Complexity Trade-off}

How large should $N_d$ be for a given $L$, traffic pattern ($p_c$) and other parameters? Theorem 1 shows that reducing $N_d$, i.e., the communication/computation complexity, results in a simpler local model but a larger error bound. Therefore, a trade-off between the optimality and complexity needs to be explored.\\

\indent
Corollary 1: {\it For a given performance $\epsilon$, 
if $N_d$ $\ge$  ${\log \left({2c_{new} {1-\epsilon_B  \over 1+\epsilon_B}L(L-N_d) \over \epsilon} \right) \over \log \left({1 \over p_c} \right) }$, 
then $E(\Delta)$ $\le$ $\epsilon$. 
For $L$ large, $L \gg N_d$ and $\epsilon_B   >0$ small, 
the condition reduces to $N_d > \Omega \left( { \log ({L \over  \epsilon}) \over  \log ({1 \over p_c}) } \right)$. \\
}

\indent
The proof of Corollary 1 can be obtained directly from Theorem 1 by letting $E[\Delta] < \epsilon$ and simple algebraic manipulations, and thus omitted. 

The corollary implies that a sufficient condition for distributed preemption decisions to be near optimal is for $N_d$ to be of an order $\log L$,  when the flow continuity probability decays exponentially with the number of hops. Meanwhile, the larger $p_c$ and the smaller $\epsilon$ are, the larger $N_d$ is. This shows clearly a trade-off between the performance and complexity.

\section{Simulations}
\label{PE}

We now study further how the performance of distributed decisions varies with respect to neighborhood size and traffic patterns through simulation. We also compare the distributed preemption with the methods used in the prior work.

\subsection{Performance Metrics and Simulation Setting}

We use the average preempted bandwidth per link at a chosen path, ${1 \over L} \sum_{k} B^k d^k$, as the performance metric in our simulation. The metric is used to quantify the effectiveness of distributed preemption in bandwidth savings. 

Our simulation generates a network topology, a certain traffic pattern, and a chosen route. To be specific, our simulation studies consider a network with two service-classes. The capacity $C$ of each link is $100$ Mbps. The bandwidths of class $1$ and $2$ flows are uniformly distributed between $1.25$ and $2.5$ Mbps, and  $2.5$ and $37.5$ Mbps, respectively. 

We use both mesh and power-law topologies in the simulations. The power law topology has 80 nodes generated through BRITE \cite{Brite}\cite{Bu}. 
The mesh topology is generated  as a planar lattice topology with 100 nodes. The nodal degree of a lattice topology is $d_0=4$ except the edge nodes. This results in flow-continuity probability $p_c$ = ${1 \over 3}$. 

The flows of each service class are evenly distributed over the network. The arrival and the departure flows of each class follow a Poisson distribution with arrival rate ($\lambda_{i}$) and departure rate ($\mu_{i}$), for $i=1,2$. 

The source and destination of a new connection are chosen at random in a network. The resulting route has about $10$ hops on average for the S-D pair of each active flow. 

We conduct over 10 experiments with random initial conditions and get the averaged values as results. For each run, active flows are routed over the shortest-path between S-D pairs and accepted if bandwidths are available, so the network is heavily populated with active flows. The distributed algorithm is used to obtain a set of local decisions. The preemption decision for the flows on the path is then obtained according to Section \ref{DA}. The performance metrics are averaged over all runs.

\subsection{Performance and Neighborhood Size}
 
A new connection setup assumes bandwidth demand $c_{new}$=$20$ Mbps and class $i_{new}=2$. Distributed preemption decisions are made by (\ref{table1}) changing  neighborhood size $1 \le N_d \le 2$. Decentralized decisions are implemented using Min-Conn algorithm \cite{Peyravian0} for comparisons. 
 There, each node makes decentralized preemption decisions independently without any cooperation with neighbors. 
The performance in the average preempted bandwidth per link is shown in Table \ref{Lat1} and \ref{PL1} for distributed preemption of different neighborhood sizes and decentralized preemption. For both topologies, the preemption costs are reduced sharply with the cooperation with neighbors.

For planar lattice topologies, the nodal degree $d_0$=4 results in the flow-continuity probability $p_c$=$1/3$. The link-dependency probability $P_{ij}$ thus decreases in $O(p_c^{|j-i|})$. As shown in Table \ref{Lat1}, even with the smallest neighborhood $N_d=1$, the preemption cost of distributed decision can be reduced by $50\%$ compared to the decentralized preemption algorithm. 

For power-law topologies, nodes have different degrees, and the path-length of a connection is around 2 or 3 hops on the average. Thus, the link-dependency probability $P_{ij}$ also decays sharply. The effectiveness of the distributed preemption is similar to that of planar lattice topologies.

\begin{table}[h!]
\caption{Preemption Costs on a Planar Lattice Topology of $d_0$=4} 
\centerline{
\begin{tabular}{|c|c|c|c|} \hline 
 Decentralized Preemption &  $N_d$=1 & $N_d$=2 \\ \hline \hline
  16.7& 7.6& 6.3 \\ \hline 
\end{tabular}}
\label{Lat1}
\end{table}

\begin{table}[h!]
\caption{Preemption Costs on a Power-Law Topology} 
\centerline{
\begin{tabular}{|c|c|c|c|} \hline 
Decentralized Preemption & $N_d$=1 & $N_d$=2 \\ \hline \hline
  17.2&  8.8& 7.5 \\ \hline 
\end{tabular}}
\label{PL1}
\end{table}

\subsection{Neighborhood Size and Traffic Patterns}

We now study how the performance of the distributed preemption varies with both $N_d$ and flow-continuity probability $p_c$. At each experiment, active flows are generated randomly for each flow-continuity probability $p_c$ and neighborhood size $N_d$. This is repeated for a wide range of $p_c$ and $N_d$ values. The preempting route has $L$ = 10 hops, and a new connection has bandwidth demand $c_{new}$=$20$ Mbps. 

\begin{figure} [htb] \centering
  \begin{tabular}{c}
    \resizebox{3.5in}{!}{\includegraphics{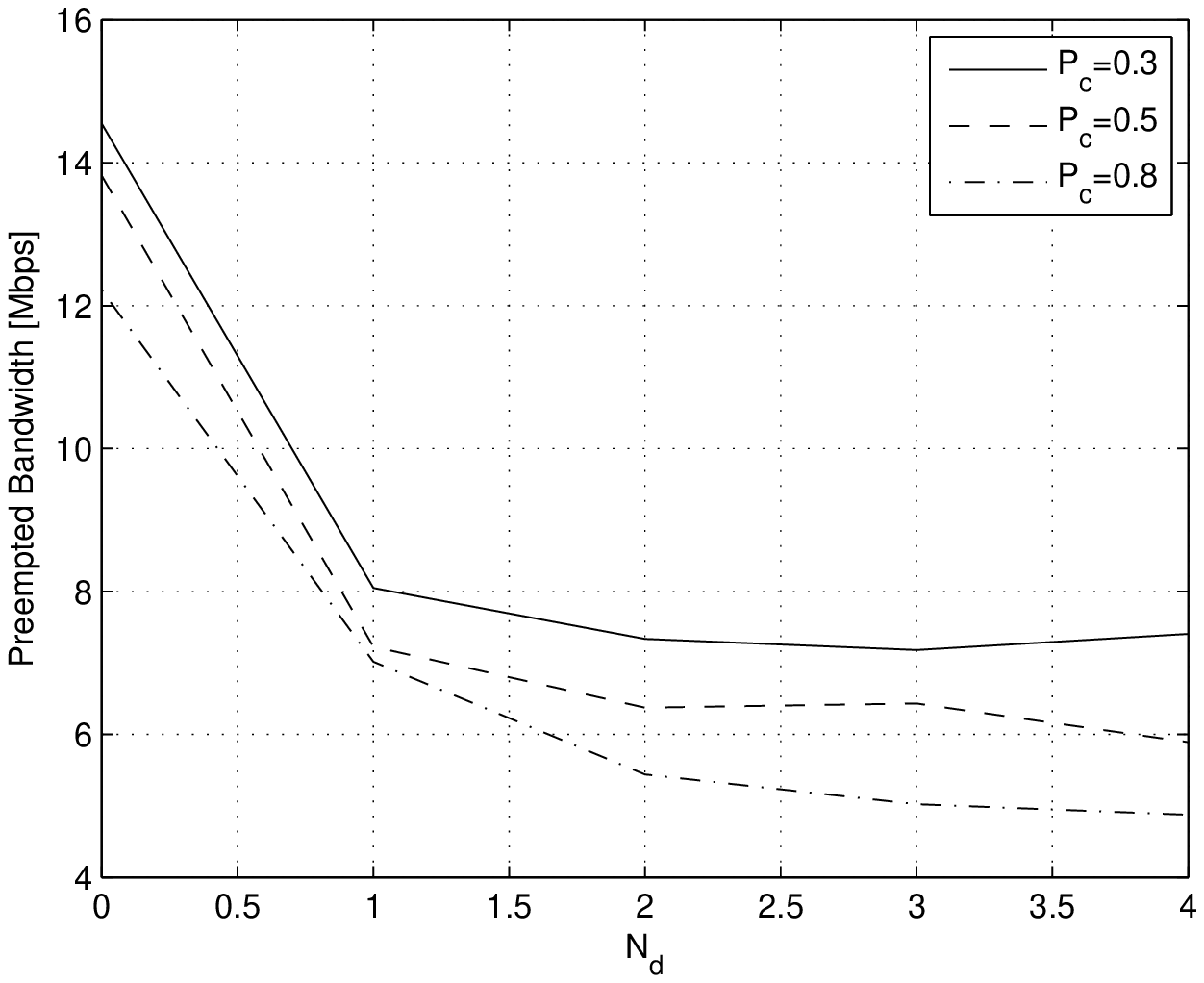}} \\ 
    {\footnotesize (a) Distributed preemption: Varying $N_d$ for different $p_c$ }\\
    \resizebox{3.5in}{!}{\includegraphics{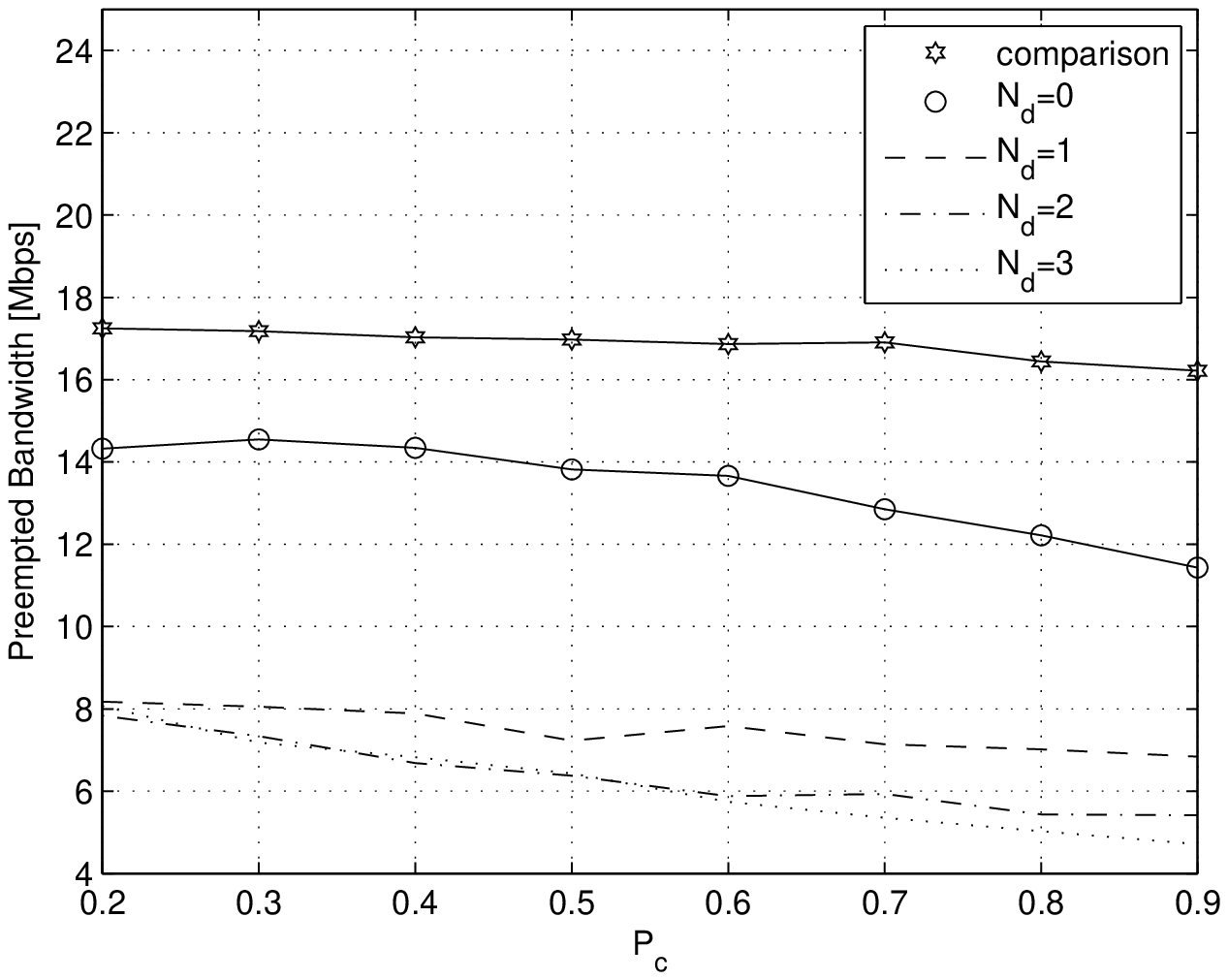}} \\
    {\footnotesize (b) Distributed preemption, compared with decentralized algorithm in \cite{Peyravian0}.}
   \end{tabular}
\caption{Average preempted bandwidth, with $c_{new}$=20 Mbps, link capacity $C$=100 Mbps, and $L$=$10$ hops on the preempting route} 
\label{N_dPB}
\end{figure}

Figure \ref{N_dPB} (a) shows that the preempted bandwidth decreases sharply by including the information only from the nearest neighbors. This is especially significant for a small $p_c$ (e.g., $p_c$=$0.3$), which corresponds to short flows. $N_d$=0 corresponds to decentralized decisions where there is no information exchange with neighbors. Hence the figure shows that the cooperation with the nearest neighbors (i.e., $N_d$=1) can improve the performance by $53 \%$. 

The cooperation with farther neighbors (e.g., $N_d$=4) results in another $3.3 \%$ bandwidth saving  for $p_c=0.3$. But the improvement is not significant given the increase of communication complexity. Hence, for short flows, the information exchange between the nearest neighbors seems to be sufficient to achieve the 
near-optimality.  

As $p_c$ increases, the dependence among decision variables on different links increases, and the performance gains are more pronounced with a larger neighborhood size. Figure \ref{N_dPB} (b) shows that the preempted bandwidth decreases linearly with an increase of $p_c$ for a given $N_d$. 
This is because that the dependence of two links increases along with $p_c$. Thus, for a given $N_d$, the amount of preempted bandwidth decreases with an increase of $p_c$. 

Figure \ref{N_dPB} (b) also shows that the preempted bandwidth of the distributed preemption is smaller than that of Min-Conn \cite{Peyravian0} algorithm. The complexity of Min-BW algorithm is $O(f_{max} \cdot 2^{f_{max}})$, which is computationally intractable for $ f_{max}$ large. Thus, Min-Conn algorithm (whose complexity is $O(f_{max}^2)$) is used for comparison.\\
%
%

%

\subsection{Path Length}

\begin{figure} [htb] \centering
  \begin{tabular}{c}
    \resizebox{3.5in}{!}{\includegraphics{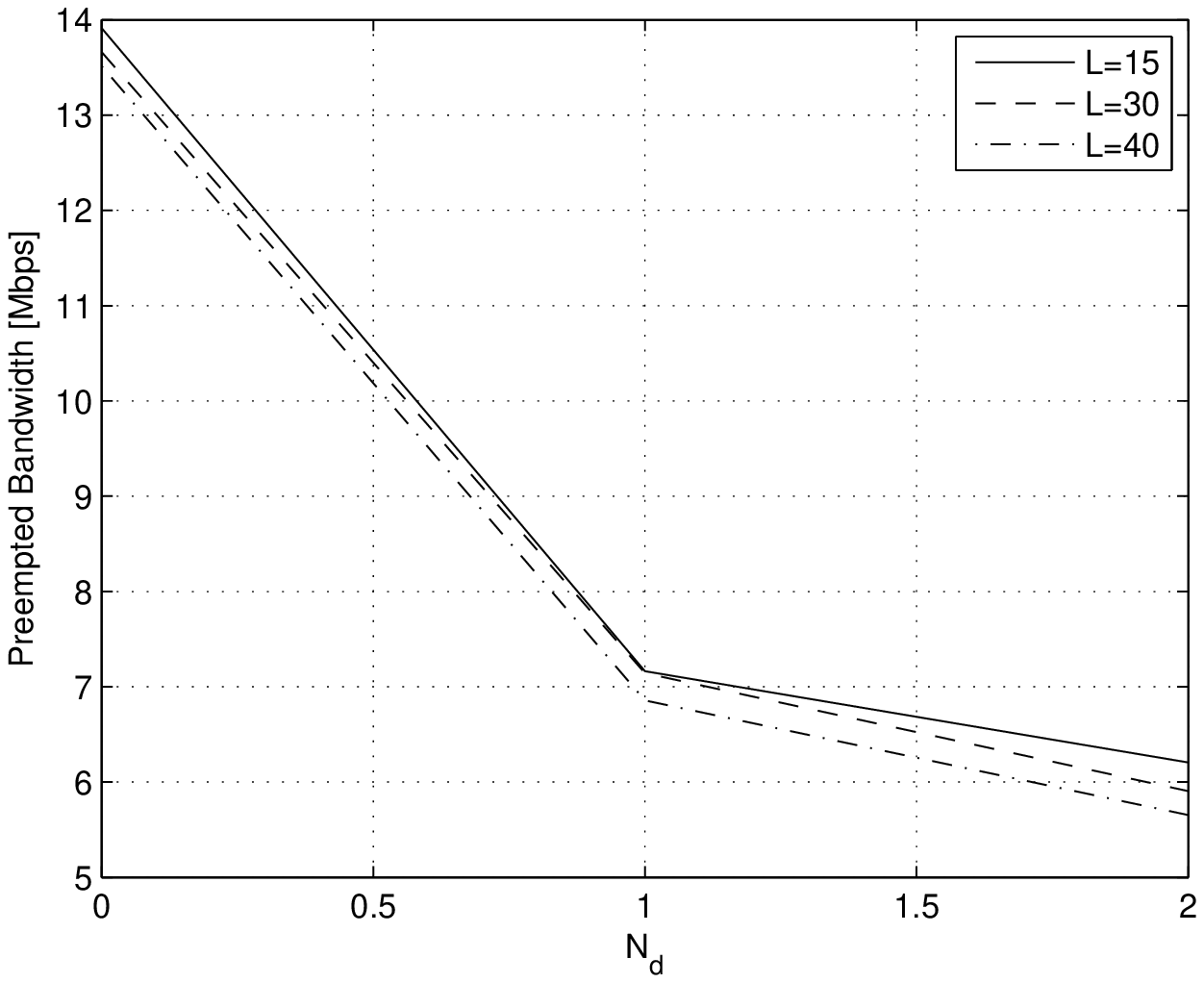}}\\ 
    {\footnotesize (a) Distributed preemption: Varying $N_d$}\\
    \resizebox{3.5in}{!}{\includegraphics{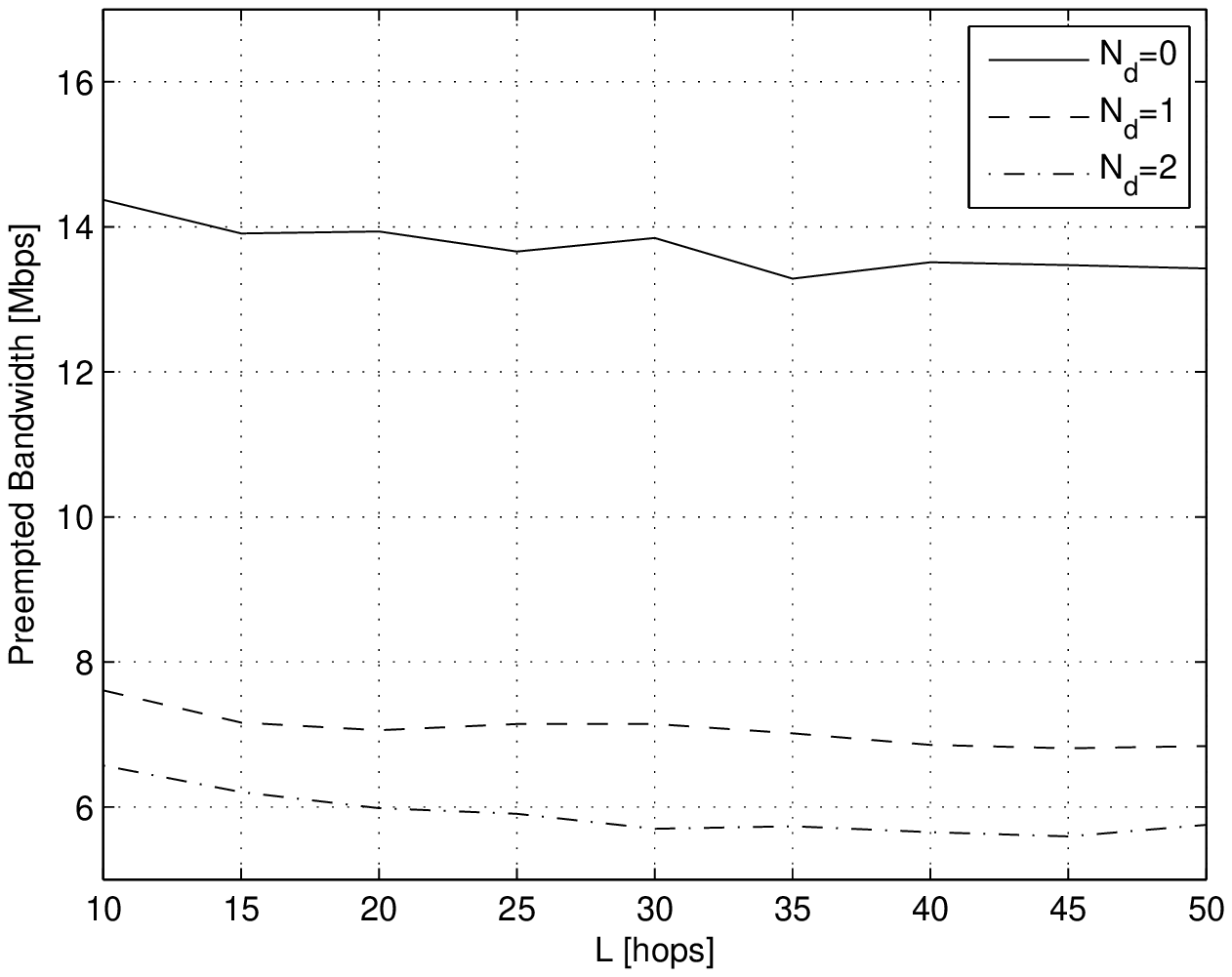}}\\ 
    {\footnotesize (b) Comparison of distributed and decentralized preemption: Varying $L$}
   \end{tabular}
\caption{Average preempted bandwidth. $c_{new}$=20 Mbps, $C$=100 Mbps, and $p_c$=0.4} 
\label{L_PB}
\end{figure}

Now we consider the impact of path length $L$ for fixed $p_c$=$0.4$.  Other parameters used are $c_{new}$=$20$ Mbps, and $C$ = $100$ Mbps. For a given $N_d$, as $L$ increases, the preemption cost decreases. 

Figure \ref{L_PB} (a) shows that for all $N_d$ values, the corresponding preempted bandwidth decreases as $L$ increases. However, the decrease of preempted bandwidth is lower bounded for $L$ $>$ $30$ hops, such as Figure \ref{L_PB} (b).


%

\subsection{Bandwidth Demand}

\begin{figure} [htb] \centering
  \begin{tabular}{c}
    \resizebox{3.2in}{!}{\includegraphics{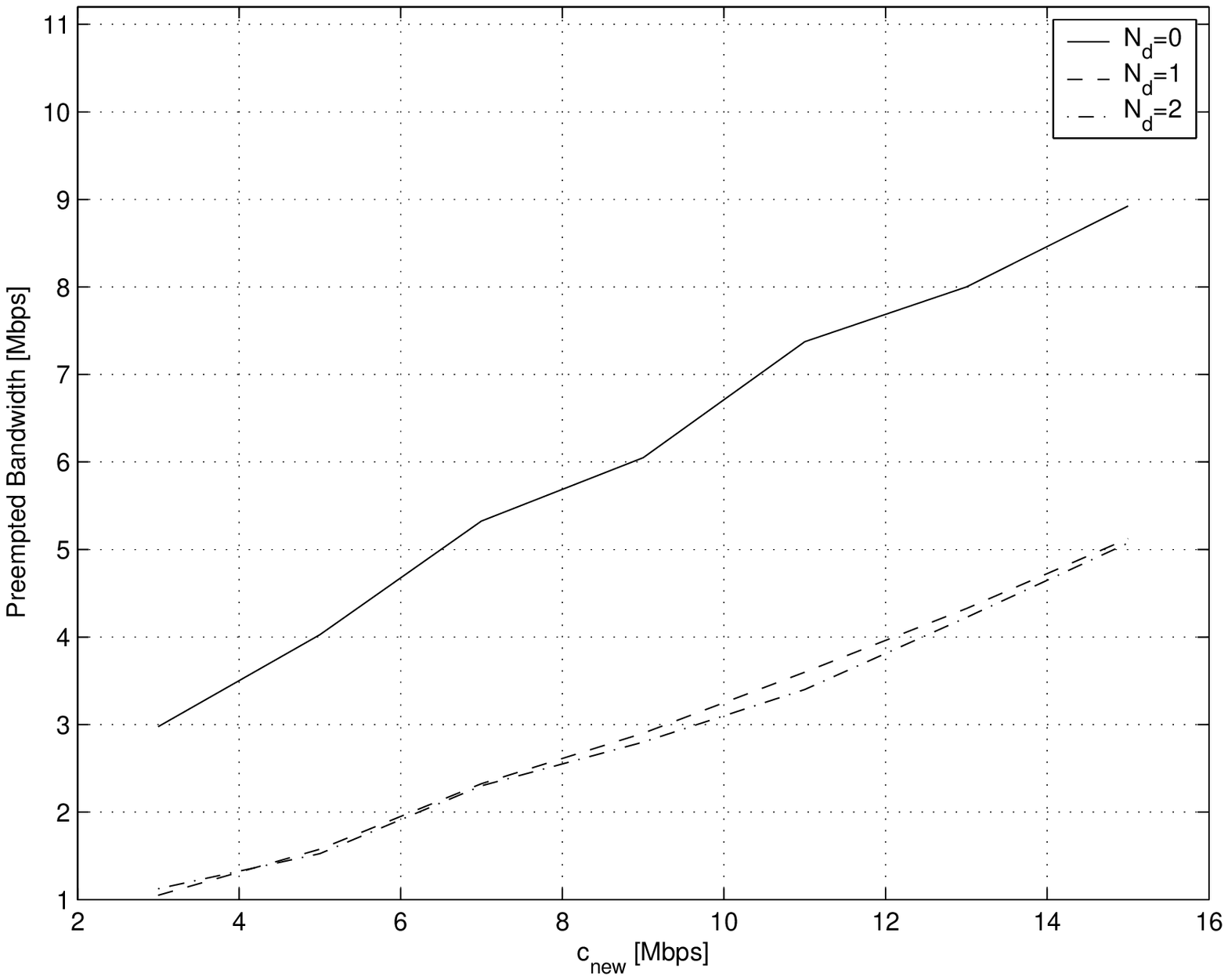}}
   \end{tabular}
\caption{Comparison of distributed and decentralized preemption} 
\label{cnew_PB}
\end{figure}

Now we consider the impact of bandwidth demand $c_{new}$ of a new connection together with the neighborhood size. Other parameters are fixed and chosen as $p_c$=$0.4$, and $C$ = $100$ Mbps. 

Figure \ref{cnew_PB} shows that for all $N_d$ values, the corresponding preempted bandwidth increases with $c_{new}$ linearly. Moreover, the gain of the distributed preemption is significant compared with decentralized preemption, and increases linearly with $c_{new}$ also.

\section{Related Work and Discussions} 
\label{Priorwork}

\indent
{\bf Connection preemption:} Garay and Gopal \cite{Garay} show that centralized connection preemption is NP-complete, and thus develop heuristic algorithms. The algorithms consider all connections on a preempting route but are not optimal in preemption decisions. 

Peyravian and Kshemkalyani \cite{Peyravian0} propose two practical decentralized algorithms. The algorithms incorporate multiple factors such as the priority of each connection, the bandwidth and the number of connections to be preempted. The algorithms are locally optimal at each link but not globally optimal at the entire route since there is no information exchange among links.
Oliveira \cite{Oliveira} formulates connection preemption through linear programming and proposes adaptive heuristic algorithms. The lower-priority connections are then preempted to minimize the impact to the active bandwidth. The optimality is not considered there. 

Stanisic \cite{Stanisic} develop two randomized preemption algorithms whose complexity is linear in the number of lower priority flows. These algorithms do not consider the optimality issues. 
Meyer et.al. \cite{Meyer} considers soft preemption that utilizes a preemption pending flag to mitigate gracefully the re-route process of preempted LSP.

As the existing algorithms provide important empirical results, the issues of optimality and complexity have been studied insufficiently. Most of these algorithms do not consider information exchange among neighbors.\\
\indent
{\bf Distributed Management:} Flow preemption can be cast in a general context of distributed management. Distributed algorithms and protocols have been designed using local information on configuration management 
(see \cite{Wattenhofer} and references therein). 
Self-organizing protocols are proposed for sensor networks and developed for p2p self-stabilizing using graph coloring (see \cite{Ko} and references therein). The distributed management, however, has been traditionally done based on heuristic local rules. Near-optimal distributed algorithms remain to be an open problem when spatial dependence is non-Markovian.\\
\indent
{\bf Probabilistic Graphical Models:} Probabilistic graphical models have been used to represent the spatial dependence in complex systems. Markov Random Fields have been used to model the cooperation of mobile agents \cite{Baras} in addition to their wide applications in image processing \cite{Geman}. Factor graphs are used to represent the dependence and as computationally efficient distributed algorithms \cite{Kschischang}.  

One related work is a cross-layer graphical model developed for optical networks \cite{Liu07}, where a Markov Random Field models the spatial dependence of routes at the network layer. The other related work is our prior work where the probabilistic graphical models are used for the distributed management of  wireless ad hoc networks \cite{JeonJi}. The work presented in this paper, however, considers a different problem of flows in a different setting of wireline networks.\\
\indent
{\bf Optimality and Complexity:} Optimality in terms of network capacity has been investigated for wireless ad hoc and sensor networks \cite{Shakkottai}. 
However, near-optimal algorithms have been studied little for controlling flows. 

\section{Conclusions}
\label{Conclusions}

In this work, we have studied distributed connection preemption in multi-class networks. The work is motivated by the fact that connection preemption is known to be NP-complete. Centralized preemption can achieve an optimal performance but has an intractable communication complexity. Decentralized preemption is computationally feasible but lack of a good performance. 
This work has focused on whether a near-optimal performance can be achieved by distributed preemption at a moderate communication complexity.  We have developed a distributed framework, where nodes make local preemption decisions through cooperation with neighbors. The framework treats distributed preemption as a machine learning problem where a large number of statistically dependent decisions can be treated jointly.  

Specifically, we have developed a probabilistic spatial model of distributed preemption decisions. We have shown that a sufficient condition for distributed preemption to be near-optimal is that the spatial model is a Markov Random Field. We have then identified a cause of spatial dependence which is due to flows trespassing multiple links. We have examined commonly-used traffic patterns including short-range dependence flows and shortest-path flows. This results in a certain sufficient conditions on the near-optimality. In particular, the sufficient conditions quantify joint impacts of the flow-continuity probability, the bandwidth demand of a new flow, the communication complexity of distributed algorithms, and route lengths for short-range dependent active flows. 

We have shown that the spatial dependence can be characterized by probabilistic graphical models. The graphical models allow us to apply  distributed algorithms based on stochastic relaxation.  We have shown through analysis and simulations that for short-range dependent flows, information exchange with only the nearest neighbors can significantly improve the performance of preemption. The use of more neighbors result in further improvements but not as pronounced given an increase in the communication complexity.

More general traffic patterns require a further study. It remains open how to quantify the validity/invalidity of distributed preemption for long-range dependent flows. For example, the long-range spatial dependence cannot be ignored in linear or ring topologies, and thus the neighborhood for distributed preemption is not localized. Hence, distributed algorithms need to be extended to multiple routes and other traffic patterns in a more realistic network setting where the impact of topologies may become significant. From a computational standpoint, one disadvantage of distributed preemption using stochastic relaxation is a slow convergence. Hence future work also involves a study of computation time in terms of delays.\\  

\indent
\begin{Acknowledgement}
Supports from NSF ECS 9908578 and Georgia Tech Broadband Institute are gratefully acknowledged. 
The authors would like to thank Guanglei Liu and Joonbeom Kim for many useful discussions.
\end{Acknowledgement}

\section*{Appendix}
\setcounter{section}{0}

\section{Proof of Lemma 1}
\label{Lemma1}

Consider two links $i$ and $m$. Let $|i-m|$ denote the distance of these two links. The distance of two links indicates the hop-counts of the shortest path between them. 

When $|i-m|$ $\le h_0$, there can be a flow that shares both links. Otherwise, i.e., $|i-m|$ $> h_0$, there cannot be any flow that shares both links. Therefore, the decisions at link $i$ are conditionally independent of the other links beyond $h_0$ hops.

\section{Proof of Lemma 2}
\label{Lemma2}

As given in assumption (1), there are multiple routes from node $i$ to node $j$ on the planar regular topology, one of which is the preempting route. Since each node has nodal degree $d_0$, the probability is  $({1 \over d_0-1})$ for a connection that passes link ($i-1$) on the preempting route to trespasses through the next link $i$. Thus, the probability that an active flow trespasses links $i-1$, $\cdots$, $j$ on the preempting route is $({1 \over d_0-1})^{|j-i|}$. 

Assume that link $i-1$ is the first link that an active flow encounters the preempting route. Consider the $k$-th link from the source node of an active flow. Assume that an active flow first meets each link on the preemption route equally likely. Then the probability that link $i-1$ on the preemption router is the $k$-th link of an active flow is ${1 \over L}$ for $1 \le k \le L$. Thus, the probability that an active flow meet on link $i-1$ at the preemption route proceeds to link $j$ is $({L-|j-i| \over L})$, where $|j-i|$ is the number of hops between link $i-1$ and link $j$. 
Therefore, a lower-bound of $P_{ij}$ is $({L-|j-i| \over L}) ({1 \over d_0-1})^{|j-i|}$.

\section{Proof of Lemma 3}
\label{Lemma3}

Consider two nodes $i$ and $j$ on the preempting route. As each connection is assumed to follows the shortest path between a source-destination 
pair, we obtain an upper bound of the probability $P_{ij}$ by counting the total number of shortest paths between node $i$ and $j$ with $|j-i|$ hops.

The total number of shortest paths from node $i$ to $j$ with $|j-i|$ hops is always upper bounded by $ C (|j-i| , {|j-i| \over 2} )$. To be specific, a shortest-path from node $i$ to $j$ is composed of $k$ horizontal and $|j-i|-k$ vertical hops, and the total number of shortest paths from node $i$ to $j$ is $C \left (|j-i| , k \right )$, for $1 \le k \le |j-i|$. Evidently, $C (|j-i| , {|j-i| \over 2} )$= max\{$ C\left (|j-i| , k \right )$, for $1 \le k \le |j-i|$\}. Thus, an upper bound of the number of shortest paths from node $i$ to $j$ with $|j-i|$ hops is $ C (|j-i| , {|j-i| \over 2} )$.

Now, consider a set of nodes which are separated from node $i$ by $|j-i|$ hops over the shortest paths. We count the total number of shortest paths from node $i$ to this set of nodes. Starting from node $i$, we can reach one of such nodes by taking $r$ horizontal steps and $|j-i|-r$ vertical steps, for $1 \le r \le |j-i|$.
For instance, with all positive vertical and horizontal steps (i.e., not going backwards), the number of 
shortest paths with the distance of $|j-i|$ hops from node $i$ is $\sum_{r=1}^{|j-i|} C(|j-i|, r)$. From binomial formula, $\sum_{r=1}^{|j-i|} C(|j-i|, r)$ = $2^{|j-i|}-1$. However, there are four combinations about the same directions of vertical/horizontal steps. The nodes that are located on the line of radian $0$, ${\pi \over 2}$, $\pi$, and ${3\pi \over 2}$ centered at node $i$ are counted twice. Thus, a lower bound of the total number of shortest-paths from node $i$ to the set of nodes is $2(2^{|j-i|}-1)$ for $|j-i|=2$, and $3(2^{|j-i|}-1)$ for $|j-i|>2$.

An upper bound of probability $P_{ij}$ is the ratio between an upper bound of  total number of shortest paths from node $i$ to $j$, and a lower bound of  total number of shortest paths from node $i$ to the set of nodes that are $|j-i|$ hops away. Moreover, as shown in Lemma 2, the probability that an active flow on link ($i-1,i$) proceeds further on the direction of the new flow by $|j-i|$ hops is $({L-|j-i| \over L})$. Thus,
{\small
\begin{eqnarray}
P_{ij} &\le&
\left \{
\begin{array}{ll}
({L-|j-i| \over L}){ C \left (|j-i|, {|j-i| \over 2} \right) \over 2 (2^{|j-i|}-1) }, & |j-i|=2 \\
({L-|j-i| \over L}){ C \left (|j-i|, {|j-i| \over 2} \right) \over 3 (2^{|j-i|}-1) }, & |j-i|>2.
\end{array}
\right.
\end{eqnarray}
}
Now consider $|j-i| >> 1$. Using Stirling formula, $n!$ $\approx$ $\sqrt{2 \pi} \mbox{exp}(-n)n^{n+0.5}$. Thus, $ C \left (|j-i|, {|j-i| \over 2} \right) \approx {2^{|j-i|+1} \over \sqrt{2 \pi |j-i|}}$, and $P_{ij}^u$ $\approx$ $({L-|j-i| \over L}){1 \over 2 \sqrt{2 \pi |j-i|}}$.

\section{Proof of Theorem 1}
\label{Theorem1}

Consider a set of randomly generated active flows and a randomly chosen preempting route of $L$ hops. 

Then, $\Delta$= $|H(\mbox{\boldmath$d^*$}) - H(\mbox{\boldmath$\hat{d}$})|$.
To find an upper bound of $\Delta$, we have 
$|H(\mbox{\boldmath$d^*$}) - H(\mbox{\boldmath$\hat{d}$})|$ =
$| \left ( H(\mbox{\boldmath$d^*$})- H^l(\mbox{\boldmath$d^*$}) \right )
+  \left ( H^l(\mbox{\boldmath$\hat d$}) - H(\mbox{\boldmath$\hat d $}) \right )
+ \left ( H^l(\mbox{\boldmath$d^*$}) - H^l(\mbox{\boldmath$\hat d $}) \right )|$
$\le$
$| \left ( H(\mbox{\boldmath$d^*$}) - H^l(\mbox{\boldmath$d^*$}) \right )
+ \left ( H(\mbox{\boldmath$d^*$}) - H^l(\mbox{\boldmath$\hat d$}) \right )|$. 
Here the inequality holds because 
$H(\mbox{\boldmath$d^*$}) \le  H(\mbox{\boldmath$\hat d $})$ and
$H^l(\mbox{\boldmath$\hat d $}) \le  H^l(\mbox{\boldmath$d^*$})$ by definition of $\mbox{\boldmath$d^*$}$ and $\mbox{\boldmath$d^l$}$.

\indent
Assume that for the global and local optimal decisions, 
$\mbox{\boldmath$d^*$}$ and $\mbox{\boldmath$d^l$}$, the second-order consistency is achieved as shown in the constraint in $H^l()$. 
That is, $d^k_i=d^k_j$ for $|i-j| \le N_d$, and for $1 \le k \le |S_F|$. 
Since $d^k_i$'s are binary, the second-order consistency implies all orders up to $L-N_d$ consistency. 
That is, for a given link $i$. Consider a set $S_i={i_1, \cdots, i_l}$ 
that contains indices of any other links within $N_d$ hops of link $i$. 
$d^k_i=d^k_{i_1}= \cdots = d^k_{i_l}$.

Thus,
{\small
\begin{eqnarray}
&   & E(|H(\mbox{\boldmath$\hat{d}$})-H^l(\mbox{\boldmath$\hat{d}$} )|) + E(|H(\mbox{\boldmath$d^*$})-H^l(\mbox{\boldmath$d^*$})|) \nonumber \\
&\le& I_2+I_3+ \cdots +I_L, 
\end{eqnarray}
}
where
{\small
\begin{eqnarray} 
I_2 &=& 2\sum_k \left (E[\sum_{i_1} \sum_{i_2 \neq i_1} d_{i_1}^{k} d_{i_2}^{k}- \right. \nonumber \\ \nonumber
& & \left. \sum_{i_1} \sum_{i_2 \neq i_1, |i_1-i_2| \le N_d } d_{i_1}^{k} d_{i_2}^{k}] \right),
\\ \nonumber
\end{eqnarray}
}, 
$\cdots$, 
{\small
\begin{eqnarray}
I_L &=& 2\sum_k \left( E[\sum_{i_1} \cdots \sum_{i_{L-1}} d_{i_1}^k \cdots d{i_{L-1}}^k] – \right. \\ \nonumber
& & \left. -E[\sum_{i_1} \cdots \sum_{i_{L-1} \neq i_1, |i_1-i_{L-1}|\le N_d} d_{i_1}^k \cdots d_{i_{L-1}}^k] \right ) 
\end{eqnarray} 
}
\noindent
for $|i_1 - i_l| \le N_d$, for $l=2, \cdots,  L$. 

Hence, within a neighborhood $N_d$ of any given link, all links make consistent decisions. Beyond such a neighborhood, links can make different and thus incorrect preemption decisions. 
To bound the error, we let $\mbox{\boldmath$d$}_a$ be a feasible preemption decision without distinguishing whether it is globally or locally optimal. Consider an active flow $k$ on link $i$ of a preempting route. For the local model $H^l(\mbox{\boldmath$d$}_a)$ of neighborhood size $N_d$, the continuity of the active flow is neglected beyond the neighborhood $N_d$, i.e., on the links ($i+N_d+m$) or ($i-N_d-m$) for $m \ge 1$. 
Hence, the error caused by using the local model can be counted by the neglected active flows beyond the neighborhood. 

Specifically, from $I_2$, the error caused by a flow $k$ that leaves at link $i+N_d+m$ corresponds to the bandwidth $B^k$ of flow $k$, and the probability that the flow leaves at this portion of the path is $p_c^{N_d+m-1}(1-p_c)$, for $m \ge 1$. Therefore, for one active flow on the preempting route, the total expected error of ignoring the second-order terms for $\forall m \ge 1$ is less than $B^k$ $\sum_{m=1}^{L}$ $(L-N_d) p_c^{N_d+m-1}(1-p_c) \le B^k p_c^{N_d} (L-N_d)$.

Thus, for a feasible configuration $\mbox{\boldmath$d$}_a$, the expected error caused by neglecting the second-order terms (i.e., $I_2$) for all flows is upper bounded by $2{c_{new} \over B_0(1-\epsilon_B)} p_c^{N_d} B_0 (1 + \epsilon_B) (L-N_d)$, where ${c_{new} \over B_0(1-\epsilon_B)}$ denotes the maximum number of flows feasible to be preempted at a link. 

Similarly, from the third term (i.e., $I_3$), the expected error caused by a flow $k$ that shares at least three links ($i$, $j$ and $l$) with the preempting route is 
$B^k (1-p_c)^2p_c^{|j-i|+|l-i|-2}$ for $|j-i|>N_d$ and $|l-i|>N_d$,  
where $(1-p_c)p_c^{|j-i|-1}$ is the probability a flow continues to the link $j$ from link $i$ and then exists the route. 
Since there are at most ${L-N_d \choose 2}$ such terms for each flow in the third term of (\ref{localmodel}), 
this error is upper bounded by $2{L-N_d \choose 2} {c_{new} (1+\epsilon_B) \over (1-\epsilon_B)} \cdot p_c^{2N_d} \cdot (L-N_d)$. 

A similar bound can be obtained for the $m$-th-order term of (\ref{obj}), for $m=2, \cdots, L$.  Thus, let $A=2c_{new} {1+\epsilon_{B} \over 1 -\epsilon_{B}}$, then
{\small
\begin{eqnarray}
E(\Delta) &\le& E(|H(\mbox{\boldmath$\hat{d}$})-
H^l(\mbox{\boldmath$\hat{d}$} )|) + E(|H(\mbox{\boldmath$d^*$})-H^l(\mbox{\boldmath$d^*$})|) \nonumber \\
&\le& A(L-N_d) \left ( {L-N_d \choose 1} p_c^{N_d} + {L-N_d \choose 2} p_c^{2N_d} + \cdots \right. \nonumber \\
& & \left. +  p_c^{N_d(L-N_d)} \right ) \nonumber \\ 
&\le& AL \left( (1+ p_c^{N_d})^{L-N_d} -1 \right ).
\end{eqnarray}
}
For $p_c^{N_d}L=o(1)$, the bound is $AL(L-N_d)p_c^{N_d} +o(L p_c^{N_d})$.

\nocite{*}
\bibliographystyle{paper}

%
\end{document}